\newif\ifreview 
\newif\ifarxiv 
\newif\ifcamera \newcommand{\cameraready}{\cameratrue}

\cameraready

\documentclass[11pt]{article}

\ifreview \usepackage[review]{acl} \fi
\ifarxiv \usepackage[final]{acl} \fi
\ifcamera \usepackage[final]{acl} \fi

\usepackage[utf8]{inputenc}
\usepackage{lmodern}
\usepackage{times}
\usepackage{latexsym}
\usepackage[T1]{fontenc}

\usepackage{microtype}

\usepackage{inconsolata}

\usepackage{bbding}
\usepackage{graphicx}
\usepackage{amsmath}
\usepackage{amssymb}
\usepackage{booktabs}
\usepackage{amsfonts}
\usepackage{algorithm}
\usepackage{algorithmicx}
\usepackage{algpseudocode}
\usepackage{footnote}

\usepackage{fancybox}
\usepackage{tcolorbox}

\usepackage{tikz}
\usetikzlibrary{shapes, arrows, positioning}
\usepackage{xcolor}
\usepackage{comment}
\usepackage{amsthm}
\usepackage{bbm} 
\usepackage{subcaption}
\usepackage{multirow}
\usepackage{comment}
\usepackage{makecell} 
\usepackage{tabularx}
\usepackage[normalem]{ulem}
\useunder{\uline}{\ul}{}
\usepackage{booktabs}
\usepackage[switch]{lineno}

\usepackage{url}
\usepackage{xspace}
\usepackage[export]{adjustbox}

\usepackage{newunicodechar}
\newunicodechar{ℝ}{\ensuremath{\mathbb{R}}}
\newunicodechar{ℕ}{\ensuremath{\mathbb{N}}}
\newunicodechar{≤}{\ensuremath{\leq}}
\newunicodechar{∃}{\ensuremath{\exists}}
\newunicodechar{ℤ}{\ensuremath{\mathbb{Z}}}
\newunicodechar{ℚ}{\ensuremath{\mathbb{Q}}}
\newunicodechar{ℝ}{\ensuremath{\mathbb{R}}}
\newunicodechar{ℂ}{\ensuremath{\mathbb{C}}}
\newunicodechar{≥}{\ensuremath{\geq}}
\newunicodechar{∈}{\ensuremath{\in}}
\newunicodechar{∉}{\ensuremath{\notin}}
\newunicodechar{∪}{\ensuremath{\cup}}
\newunicodechar{∩}{\ensuremath{\cap}}
\newunicodechar{∅}{\ensuremath{\emptyset}}
\newunicodechar{∀}{\ensuremath{\forall}}
\newunicodechar{→}{\ensuremath{\to}}
\newunicodechar{⇒}{\ensuremath{\Rightarrow}}
\newunicodechar{±}{\ensuremath{\pm}}
\newunicodechar{×}{\ensuremath{\times}}
\newunicodechar{·}{\ensuremath{\cdot}}
\newunicodechar{−}{\ensuremath{-}}
\newunicodechar{∧}{\ensuremath{\land}}
\newunicodechar{√}{\ensuremath{\sqrt{}}}

\usepackage{color}
\usepackage{colortbl}
\definecolor{RowGray}{rgb}{0.9373,0.9373,0.9373}

\newcommand{\nbf}[1]{{\noindent \textbf{#1}}}

\newcommand{\eat}[1]{}

\newcommand{\cbit}{\begin{compactitem}}
\newcommand{\ceit}{\end{compactitem}}
\newcommand{\cben}{\begin{compactenum}}
\newcommand{\ceen}{\end{compactenum}}

\usepackage[capitalize]{cleveref}
\crefname{section}{Sec.}{Secs.}
\crefname{table}{Table}{Tables}
\crefname{figure}{Fig.}{Figs.}
\crefname{algocf}{alg.}{algs.}
\Crefname{algocf}{Algorithm}{Algorithms}

\definecolor{headercolor}{RGB}{240, 255, 240}
\definecolor{rowcolor1}{RGB}{255, 240, 240} %
\definecolor{rowcolor2}{RGB}{240, 240, 255}
\definecolor{rowcolor3}{rgb}{1,1,1}

\definecolor{pink1}{HTML}{FAD5D1}
\definecolor{pink2}{HTML}{EE9A8F}
\definecolor{pink3}{HTML}{C6464C}

\definecolor{blue1}{HTML}{A9B8ED}
\definecolor{blue2}{HTML}{7886CB}
\definecolor{blue3}{HTML}{4050B5}

\definecolor{problem-title}{HTML}{d0cece}
\definecolor{nlr-title}{HTML}{96abdd}
\definecolor{nlr-back}{HTML}{b9c6e8}
\definecolor{sr-title}{HTML}{e69a9f}
\definecolor{sr-back}{HTML}{e69a9f}
\definecolor{ar-title}{HTML}{cae4b5}
\definecolor{ar-back}{HTML}{b1d592}
\definecolor{sm-title}{HTML}{f9db6e}
\definecolor{sm-back}{HTML}{fae79c}

\usepackage{pifont}

\usepackage{fancyhdr}
\newcommand{\datasetname}{{MPM}\xspace}
\newcommand{\methodname}{{CoR}\xspace}
\newcommand{\modelname}{{CoR-Math-7B}\xspace}

\def\paperTitle{
\vspace{-1.0ex}
\begin{center}
    \makebox[\textwidth][c]{
        \raisebox{-0.2em}{\includegraphics[width=2em]{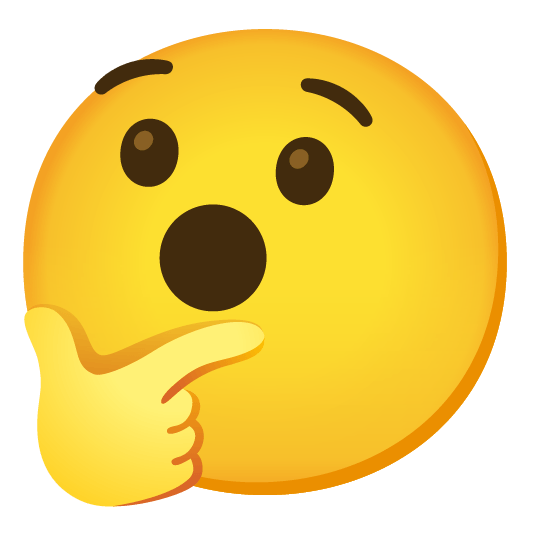}} 
        \begin{tabular}{c}
            Chain-of-Reasoning: Towards Unified Mathematical Reasoning in \\
            Large Language Models via a Multi-Paradigm Perspective
        \end{tabular}
    }
\end{center}
}

\title{\paperTitle}

\def\authorBlock{
    Yiyao Yu$^\textnormal{1}$\footnotemark[1] \qquad
    Yuxiang Zhang$^\textnormal{1}$\footnotemark[1] \qquad
    Dongdong Zhang$^\textnormal{2}$\footnotemark[2] \qquad
    Xiao Liang$^\textnormal{1}$
    \\
    \textbf{
    Hengyuan Zhang$^\textnormal{1}$ \qquad
    Xingxing Zhang$^\textnormal{2}$  \qquad
    Ziyi Yang$^\textnormal{2}$  \qquad
    Mahmoud Khademi$^\textnormal{2}$
    } \\
    \textbf{
    Hany Awadalla$^\textnormal{2}$  \qquad
    Junjie Wang$^\textnormal{1}$\footnotemark[2] \qquad
    Yujiu Yang$^\textnormal{1}$\footnotemark[2] \qquad
    Furu Wei$^\textnormal{2}$
    }
    \vspace{-0.25em}
    \\\\\fontsize{10}{10}
    \selectfont{$^\textnormal{1}$Tsinghua University} \quad
    \selectfont{$^\textnormal{2}$Microsoft} \\
    {\tt\small yuyy23@mails.tsinghua.edu.cn}, \quad
    {\tt\small joell070408@gmail.com}, \quad
    {\tt\small wangjunjie@sz.tsinghua.edu.cn}
    \\
    {\tt\small {\{dozhang,xizhang,mkhademi,hanyh,fuwei\}}@microsoft.com}
}
\author{\authorBlock}

\begin{document}
\maketitle

{
  \renewcommand{\thefootnote}%
  {\fnsymbol{footnote}}
  \footnotetext[1]{Equal contribution. Yiyao Yu did this work during the internship at Microsoft Research Asia. Yuxiang Zhang did this work as a research assistant at Tsinghua University.}
  \footnotetext[2]{Corresponding Author.}
}

\begin{abstract}
Large Language Models (LLMs) have made notable progress in mathematical reasoning, yet they often rely on single-paradigm reasoning that limits their effectiveness across diverse tasks. 
In this paper, we introduce Chain-of-Reasoning (CoR), a novel unified framework that integrates multiple reasoning paradigms \textemdash{} Natural Language Reasoning (NLR), Algorithmic Reasoning (AR), and Symbolic Reasoning (SR) \textemdash{} to enable synergistic collaboration. 
CoR generates multiple potential answers using different reasoning paradigms and synthesizes them into a coherent final solution. 
We propose a Progressive Paradigm Training (PPT) strategy that allows models to progressively master these paradigms, culminating in the development of \modelname. 
Experimental results demonstrate that \modelname significantly outperforms current SOTA models, achieving up to a $41.0\%$ absolute improvement over GPT-4o in theorem proving tasks and a $15\%$ improvement over RL-based methods on the MATH benchmark in arithmetic tasks. 
These results show the enhanced mathematical comprehensive ability of our model, enabling zero-shot generalization across tasks.
The code is available at~\url{https://github.com/microsoft/CoR}.
\end{abstract}

\section{Introduction}

\begin{figure}[!t]
\centering
\includegraphics[width=0.48\textwidth]{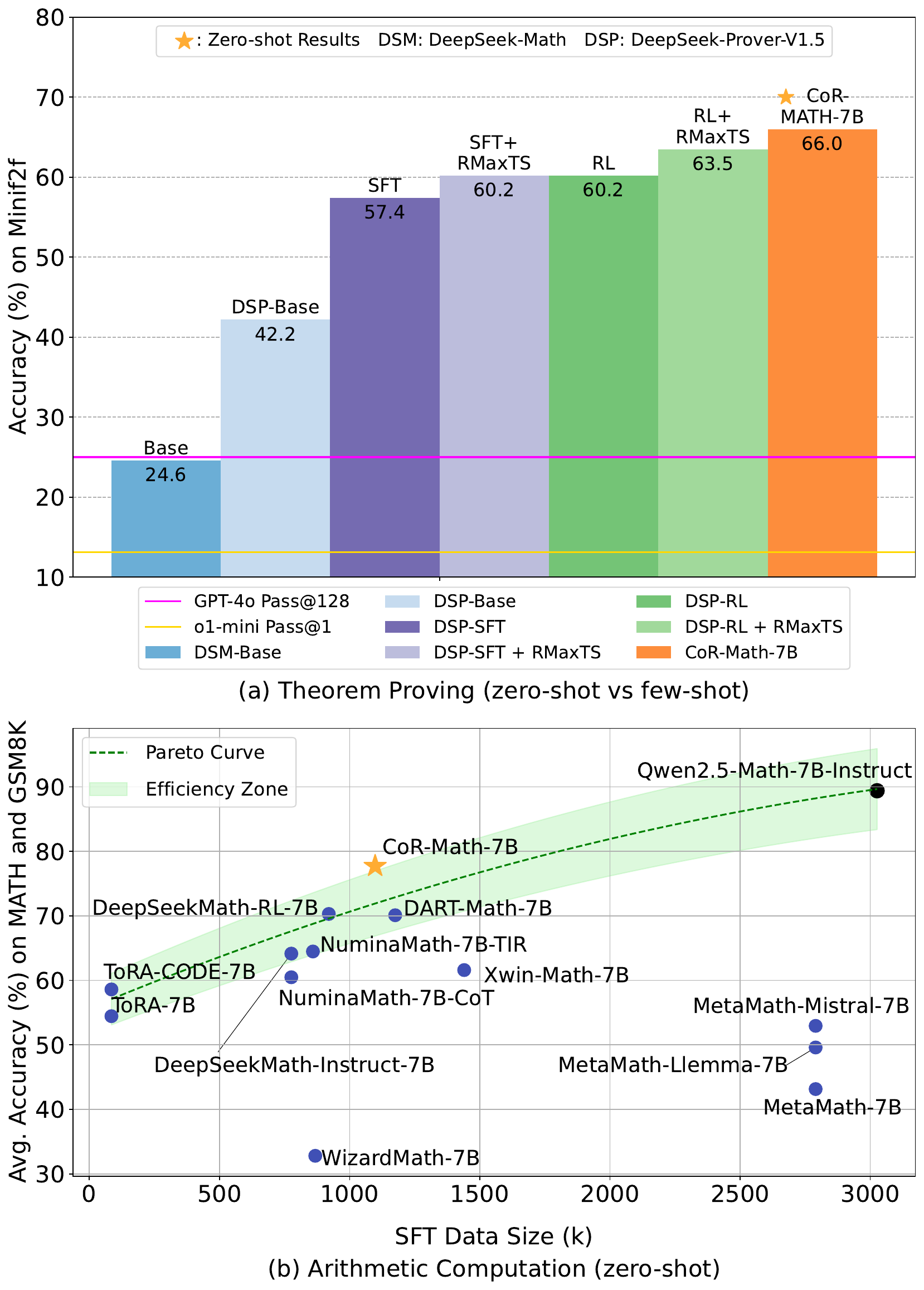}
\caption{A comprehensive comparative analysis of \modelname and baseline models across mathematical tasks. 
(a) shows the effectiveness of \modelname (zero-shot) in theorem proving tasks. 
(b) shows a resource-efficiency analysis for arithmetic computation tasks, where \modelname achieves optimal resource efficiency and near-optimal zero-shot performance.
}
\label{fig:intro-result}
\vspace{-1.25em}
\end{figure}

While LLMs have shown strong performance in solving mathematical tasks~\citep{feigenbaum1963computers,DBLP:conf/emnlp/HosseiniHEK14}, advanced open-source reasoners still struggle with solving comprehensive mathematical problems, including both arithmetic computation and theorem proving.

\begin{figure*}[!t]
\centering
\includegraphics[width=0.98\textwidth]{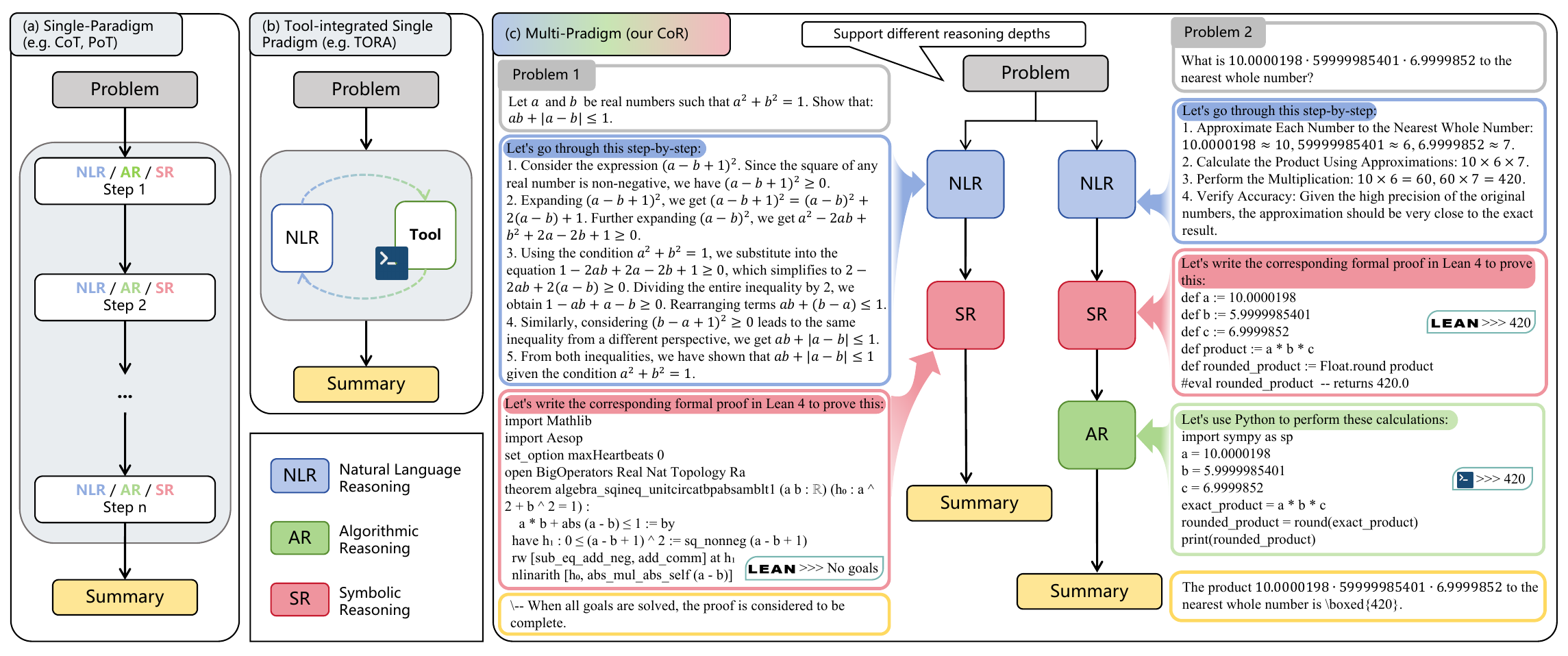}
\caption{
The reasoning process under different paradigms: (a) In single-paradigm reasoning, each reasoning step relies on the same knowledge medium, such as Natural Language (NL), algorithms, or symbols. (b) In tool-integrated single-paradigm, NL is used for reasoning, while code assists in solving specific sub-problems. After obtaining the execution results, the reasoning continues using NL. (c) The proposed CoR reasoning framework, along with several examples, shows that multi-paradigm reasoning allows for varying reasoning depths to address different types of problems.}
\label{fig:intro}
\vspace{-1.0em}
\end{figure*}

Existing works~\citep{DBLP:journals/corr/abs-2408-08152@deepseek-prover-v1.5,DBLP:journals/corr/abs-2409-12122@qwen25-math,DBLP:journals/corr/abs-2410-15700@internlm25_step,DBLP:journals/tois/ZhangWZSY24@ssr} are often trained on specific tasks, aiming to enhance their ability to independently derive answers based on specific structured knowledge representation.
This representation is known as the reasoning paradigm, involving Natural Language Reasoning (NLR), Algorithmic Reasoning (AR), and Symbolic Reasoning (SR), as depicted in~\cref{fig:intro} (a).
Specifically, NLR uses natural language for reasoning via common sense and semantic context, providing explicit step-by-step explanations~\citep{DBLP:conf/nips/Wei0SBIXCLZ22@cot}.
AR leverages code to emphasize computational operations and execution processes, such as generating Python code for execution~\citep{DBLP:journals/tmlr/ChenM0C23@pot,DBLP:conf/icml/pal}. 
SR uses logical symbols and axiomatic systems for formalized reasoning, with recent methods~\citep{DBLP:journals/corr/abs-2408-08152@deepseek-prover-v1.5,DBLP:conf/iclr/HuangLLCXWLSL24@mustard,DBLP:journals/corr/abs-2410-15700@internlm25_step} considering numerous symbolic trajectories via tree-based search for theorem proving.
However, these methods mainly focus on optimizing single-paradigm reasoning, creating models that demonstrate asymmetrical performance across different mathematical tasks. 
For instance, models specialized in NLR may exhibit deficiencies in theorem proving and vice versa. 
This fragmented paradigm constrains the upper-bound performance on individual tasks and undermines the model's capacity for cross-paradigm generalization.

Researchers have explored various strategies to tackle these challenges.
To improve the single-task performance, some works integrate tools to overcome the limitations of single-paradigm reasoning~\citep{DBLP:conf/iclr/GouSGSYHDC24@tora,numina_math_dataset@numina}, as shown in~\cref{fig:intro} (b). 
While recognizing the benefits of combining reasoning paradigms, they still rely on a single paradigm, neglecting the possibility that the second paradigm could independently complete the reasoning, thereby constraining overall potential.
Besides, to improve cross-task generalization, several studies~\citep{DBLP:journals/corr/abs-2402-03300@deepseekmath,DBLP:conf/iclr/HuangLLCXWLSL24@mustard} incorporate diverse task samples into large-scale datasets, such as those drawn from theorem proving tasks that focus exclusively on SR solutions, or from arithmetic problems that emphasize NLR solutions.
Although models trained on such data are capable of cross-task reasoning, they still rely on demonstrations for effective transfer.

To address these limitations, we propose Chain-of-Reasoning (CoR), a unified framework integrating NLR, AR, and SR to produce synergistic benefits.
As shown in~\cref{fig:intro} (c), CoR enables multi-paradigm reasoning by applying different reasoning paradigms to derive multiple potential answers, which are then summarized into a final solution.
This framework allows iterative reasoning across paradigms, leveraging prior results to enhance single-task performance. 
Moreover, CoR unifies multi-paradigm reasoning across tasks, enabling zero-shot generalization.
Adjusting prompts to control reasoning depth improves adaptability to diverse tasks.
As a result, we introduce Multi-Paradigm Mathematical (\datasetname), a dataset comprising $167$k reasoning paths, and propose Progressive Paradigm Training (PPT), a method enabling models to progressively master multiple reasoning paradigms, leading to \modelname.

We evaluate \modelname on five challenging mathematical reasoning benchmarks, covering both arithmetic computation and theorem proving.
Our results show that LLMs equipped with CoR significantly surpass current SOTA baselines.
In theorem proving (\cref{fig:intro-result} (a)), \modelname improves zero-shot performance over GPT-4o~\citep{DBLP:journals/corr/abs-2410-21276@gpt4o} by $41.0\%$, outperforming all few-shot reasoners.
For arithmetic tasks, \modelname achieves a $24.2\%$ absolute improvement over GPT-4 on MATH.
Furthermore, as shown in~\cref{fig:intro-result} (b), \modelname efficiently utilizes resources to surpass the optimal performance curve of single-paradigm approaches.
Unlike mainstream methods that conduct extensive searches within a single paradigm, \methodname enhances test-time inference in multiple paradigms.
These results show that \methodname can solve comprehensive mathematical problems through multi-paradigm reasoning, requiring less training data and lowering reasoning costs.

\section{Related Work}

\nbf{Reasoning Paradigms in LLMs.} 
Recent advancements in LLMs focus on single-paradigm reasoning, with each paradigm representing a distinct method for knowledge representation and logical inference. 
NLR uses human-readable language for commonsense reasoning and step-by-step deductions~\citep{DBLP:conf/nips/Wei0SBIXCLZ22@cot,DBLP:conf/nips/YaoYZS00N23@tot,DBLP:conf/iclr/ZhouSHWS0SCBLC23,DBLP:conf/aaai/BestaBKGPGGLNNH24,DBLP:conf/icml/SelAK0024}, AR generates executable code for precise calculations~\citep{DBLP:journals/tmlr/ChenM0C23@pot,DBLP:journals/corr/abs-2308-12950@codellama}, and SR utilizes logical symbols for theorem proving~\citep{DBLP:journals/corr/abs-2408-08152@deepseek-prover-v1.5}.
Some methods enhance reasoning by integrating external tools like calculators or code interpreters, yet they remain within a single paradigm~\citep{DBLP:conf/iclr/GouSGSYHDC24@tora}.
While effective in specific tasks, they struggle with cross-domain generalization and dynamic environments.
To address these limitations, the CoR framework enables collaboration across reasoning paradigms, enabling zero-shot multitask generalization.

\nbf{Mathematical Problem Solving with LLMs.} 
Advanced research highlights the potential of LLMs in solving mathematical problems~\citep{DBLP:conf/acl/ZhuWZZ0GZY23@core, lin2024critical,luo2025ursa}. 
Several studies have developed unified solvers by synthesizing mathematical data to address challenges like arithmetic computation and theorem proving~\citep{DBLP:conf/iclr/HuangLLCXWLSL24@mustard,DBLP:journals/corr/abs-2402-03300@deepseekmath}. 
However, most approaches focus on optimizing a single paradigm.
For instance, in arithmetic computation, tools are integrated to assist natural language reasoning.~\citep{DBLP:conf/iclr/GouSGSYHDC24@tora}. 
In theorem proving, models rely on specialized data like pre-training data and Reinforcement Learning (RL) reward data, and tree-based search methods to generate numerous possible solutions~\citep{ying2024internlmmath@internlm-math,DBLP:journals/corr/abs-2408-08152@deepseek-prover-v1.5}. 
These approaches either neglect complete reasoning or depend on large-scale search within the solution space, limiting performance gains within a single paradigm.
To address these, we introduce the \modelname model with complete reasoning processes to explore an expanded multi-paradigm solution space.

\begin{figure*}
\centering
\includegraphics[width=0.98\textwidth]{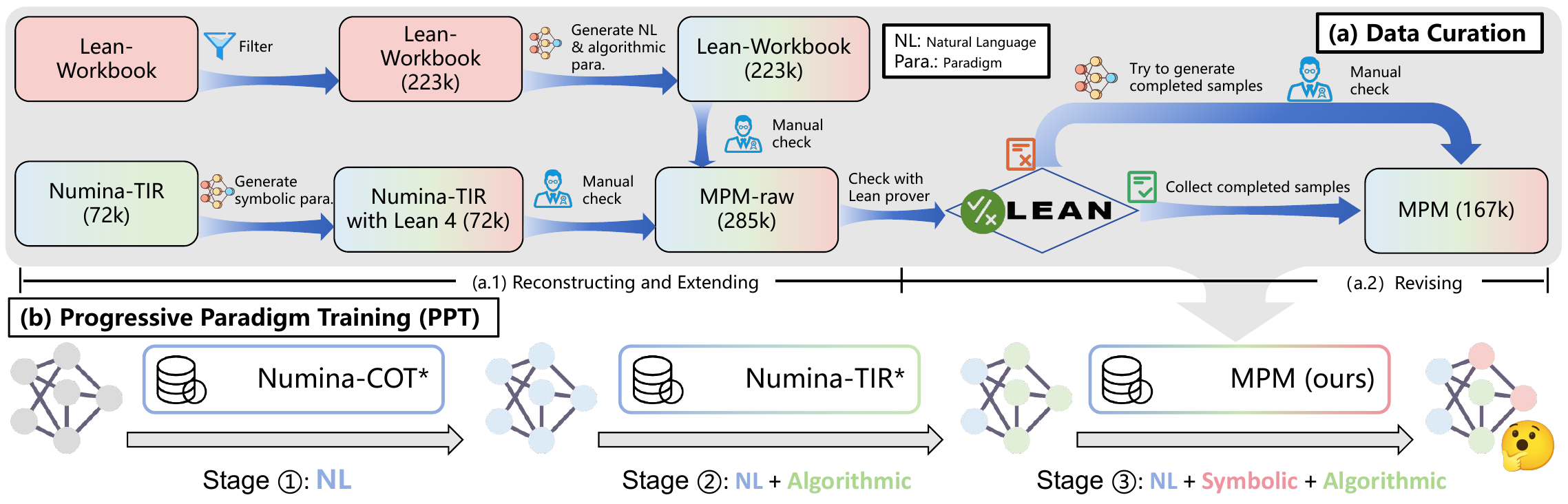}
\caption{An overview of (a) the Multi-Paradigm Math (MPM) dataset construction process, involving reconstruction, extension, and theorem prover verification, and (b) the Progressive Paradigm Training (PPT) method, where the model is trained with increasing reasoning paradigms in stages.}
\label{fig:data-train}
\vspace{2em}
\end{figure*}

\section{Chain-of-Reasoning Framework}
\label{sec:method}

\subsection{Overview}

CoR aims to enable LLMs to perform a series of multi-paradigm reasoning on any type of mathematical problem, ultimately arriving at a solution. 
Specifically, given a mathematical problem $x$, LLMs ($\mathbb{P}$) can infer the result $y$ by following multiple reasoning paradigms, where each reasoning paradigm $\tau$ includes $n$ reasoning paths $\{ rp_1, ..., rp_n \}$.
We represent this single-paradigm scenario as $y\sim\mathbb{P}(y|x,\tau)$.
To simplify the process for each reasoning paradigm, we set $n=1$ as default (Details in~\cref{ass:reasoning_hierarchy}).

Inspired by recent works~\citep{DBLP:conf/nips/Wei0SBIXCLZ22@cot} in encouraging step-by-step reasoning, we introduce CoR, which extends from a single paradigm to three paradigms $\Gamma=(\tau_1,\tau_2,\tau_3)$.
For generating multiple chained reasoning paradigms for a given problem, \methodname follows the steps outlined below.
The reasoning process begins with the problem $x$ and the first reasoning paradigm $\tau_1$. 
Subsequently, each paradigm $\tau_i$ in the sequence $\Gamma$ is generated based on the problem $x$ and the previously established paradigms $\tau_1, ..., \tau_{i-1}$, as represented by $\tau_i\sim\mathbb{P}(\tau_i|x,\tau_1,...,\tau_{i-1})$.
Finally, the outcomes of all the reasoning paradigms are aggregated to derive the final result $y$, expressed as $y\sim\mathbb{P}(y|x,\tau_{\text{NLR}},\tau_{\text{SR}},\tau_{\text{AR}})$, considering three paradigms: NLR, SR, and AR. 
In detail, $\tau_{\text{SR}}$ and $\tau_{\text{AR}}$ include both the complete reasoning paths and the interaction processes with tool outputs. 
In this study, $\tau_{\text{SR}}$ uses the Lean prover, and $\tau_{\text{AR}}$ applies a Python compiler to obtain reasoning results.

In our workflow (\cref{fig:data-train}), the training pipeline includes:
(a) collecting the \datasetname dataset with deep reasoning paths;
(b) using progressive paradigm training to enhance reasoning across paradigms; and
(c) applying the trained LLMs for zero-shot inference with sequential multi-paradigm sampling to explore diverse solutions.

\subsection{Collecting Dataset}
\label{ss:collecting_dataset}

To train \methodname model, we extend the single-paradigm datasets to incorporate multiple reasoning paradigms, denoted as $<\texttt{x}, \texttt{NLR},  \texttt{SR}, \texttt{AR}, \texttt{y}>$. 
As presented in~\cref{fig:data-train} (a), the training data collection process involves two stages: (a.1) Reconstructing and Extending, and (a.2) Revising. 
In the first stage, we reconstruct high-quality open-source mathematical data as seed samples and synthesize additional reasoning paradigms to form the \datasetname-raw dataset. 
In the second stage, a theorem prover examines \datasetname-raw samples, and a mathematical reasoner revises those that fail, compiling all completed data into the final \datasetname dataset.

\nbf{Stage 1: Reconstructing and Extending}.
We introduce a universal text template for multi-paradigm reasoning (details in~\cref{ass:universal-template}), which standardizes the placement and relationships of reasoning paradigms.
It supports various reasoning depths and flexible combinations of different reasoning paradigms. 
As shown in~\cref{fig:intro} (c), Problem 1 shows an instance incorporating both NLR and SR paradigms, and Problem 2 integrates three reasoning paradigms. 
To ensure data integrity and prevent potential biases, we pre-process the Numina-TIR~\citep{numina_math_dataset@numina} and Lean-Workbook~\citep{DBLP:journals/corr/abs-2406-03847@lean-workbook} datasets in two steps. 
First, samples without corresponding solutions are removed. 
Second, we further reconstruct and extend these datasets by leveraging powerful LLMs $\mathcal{G}$, such as GPT-4o~\citep{DBLP:journals/corr/abs-2410-21276@gpt4o}. 
These models generate missing reasoning paradigms $\tau_{g}$ and refine existing ones $\tau^{\prime}$, ensuring comprehensive coverage and logical consistency. 
To effectively guide the processes of augmentation and refinement, we develop tailored prompts $p_{s}$ for each seed dataset (examples in~\cref{ass:enhance-prompt}).
The procedure can be described in:
\begin{equation}
\tau_{g} \sim \mathbb{P}_{\mathcal{G}}(\tau_{g} \mid p_{s} \oplus x \oplus y \oplus \tau^{\prime}), 
\tau^{\prime} \in \{ \tau_{\text{NLR}}, \tau_{\text{SR}}, \tau_{\text{AR}} \},
\end{equation}
where $\oplus$ means concatenation.
After that, we conduct a manual review of all samples.
Given that alternative approaches are readily verified through external tools, we focus on the accuracy of the NLR and AR. 
This approach considerably lowers the requisite skill level of annotators. 
To prevent data leakage, we compute the Levenshtein distance~\citep{10.5555/1822502@levenshtein} between training and test problems, as detailed in~\cref{ass:data-leak}.
As a result, this phase yields the \datasetname-raw dataset, comprising approximately $285,000$ synthetic samples.

\nbf{Stage 2: Revising}.
The \datasetname-raw dataset interacts with the Lean prover to verify proof steps, guiding the filtering and modification of reasoning paths. 
The proof $\tau_{\text{SR}}$ is submitted to the prover, and if verification is successful without errors, the entire multi-paradigm reasoning path is collected to the \datasetname dataset.
Otherwise, the error information $\varepsilon$ returned by the prover is fed into a revising model $\mathbb{P}_{R}$. 
Furthermore, this model generates a revised proof $\tilde{\tau}_{\text{SR}}$ based on a prompt $p_{\varepsilon}$. 
The relationship can be expressed as: 
\begin{equation}
\tilde{\tau}_{\text{SR}} \sim \mathbb{P}_{R}(\tilde{\tau}_{\text{SR}} \mid p_{\varepsilon} \oplus x \oplus y \oplus \tau_{\text{SR}}), 
\end{equation}
where the revised proof $\tilde{\tau}_{\text{SR}}$ is then resubmitted to the Lean prover for verification. 
This iterative process continues for up to 64 iterations or until $\tilde{\tau}_{\text{SR}}$ is verified as correct.
In detail, we employ DeepSeek-Prover-V1.5~\citep{DBLP:journals/corr/abs-2408-08152@deepseek-prover-v1.5} as the $\mathbb{P}_{R}$.

Consequently, the \datasetname dataset comprises $82,770$ problems and $167,412$ multi-paradigm reasoning solutions.

\subsection{Training}

Inspired by recent advances~\citep{numina_math_dataset@numina}, we introduce the Progressive Paradigm Training (PPT) strategy, enabling LLMs to gradually master diverse reasoning paradigms. 
As shown in~\cref{fig:data-train} (b), this framework expands the model's reasoning abilities by sequentially introducing different paradigms at each stage.
Each training stage uses a different combination of reasoning paradigms.
In stage \ding{192}, given the dominance of NL in the pre-training data of language models~\citep{DBLP:journals/corr/abs-2407-21783@llama3}, we create Numina-CoT$^*$ as an initialized teaching stage.
This is done by modifying the original Numina-CoT dataset~\citep{numina_math_dataset@numina} according to our universal text template, enabling the model to learn to use NL to solve complex mathematical problems.
Based on the question $x$, the model performs reasoning $\tau_{\text{NLR}}$ and generates the answer $y$. 
The generated sequence is $z=[x]\tau_{\text{NLR}}y$, where $[x]$ represents the inputs after tokenization.
For simplicity, we first consider the loss function for a single sample:
\begin{equation}
\mathcal{L}_{\text{sample}} = - \sum_{t=1}^{|z|} \log \mathbb{P}_\theta(z_t \mid z_{<t}),
\end{equation}
where $\theta$ represents the model parameters, $z_t$ is the $t$-th token in the generated sequence, and $z_{<t}$ indicates all tokens before the $t$-th token in the generated sequence.
In stage \ding{193}, considering a certain proportion of code corpora in the pre-training data, we expand the training dataset to include two paradigms: NLR and AR. 
Similar to Numina-CoT$^*$, we modify the original Numina-TIR to create Numina-TIR$^*$.
With this modified dataset, the generated sequence is $z=[x]\tau_{\text{NLR}}\tau_{\text{AR}}y$.
After this stage, the model can handle problems that require precise answers. 
In stage \ding{194}, we further expand the training data to three reasoning paradigms by utilizing the \datasetname dataset, where $z$ is $[x]\tau_{\text{NLR}}\tau_{\text{AR}}\tau_{\text{SR}}y$.
After full stages, the trained \modelname model not only masters NLR and AR but also can perform rigorous logical SR.

Unlike traditional curriculum and incremental learning, which sequencing tasks by increasing difficulty within a single paradigm or accumulating knowledge, PPT progressively integrates fundamentally distinct reasoning paradigms, transitioning from familiar to unfamiliar, while fostering synergy across paradigms.

\subsection{Inference}
\label{ss:inference}

We present an inference method combining variable reasoning depth and multi-paradigm sampling to solve comprehensive mathematical tasks.

\nbf{Prompts with variable reasoning depth.}
In the zero-shot inference phase, \modelname exhibits proficiency in multi-paradigm reasoning, enabling adjustable reasoning depths by modifying prompts based on task requirements.
Initially, the model is prompted to conduct NLR, thereby activating a wide range of knowledge patterns associated with natural language from the pre-training corpus. 
Subsequently, the inference method is tailored to the specific problem type. 
For example, in theorem proving, the model switches to SR, which is more suitable for formal deduction.
Since the SR output is structured in Lean 4, the relevant proof segment can be extracted as the final solution.
In arithmetic computations, the model first employs NLR, followed by SR for logical coherence, and concludes with AR for precise calculations.
A summary box is utilized to present the final results. 
The paradigm-switching behavior depends on the presence of prompts and can be categorized into instruction-followed reasoning and instruction-free reasoning. This section primarily presents results from the former. Instruction-free reasoning, which explores how the model autonomously switches paradigms without prompt guidance, will be demonstrated in ~\cref{ass:instruction-free-reasoning-case}.
This adaptable paradigm effectively captures the distinct reasoning patterns inherent in different paradigms and demonstrates flexibility in accommodating a variety of scenarios.

\nbf{Sequential Multi-Paradigm Sampling (SMPS).}
Instead of token-level sampling based on tree structures~\citep{DBLP:journals/corr/abs-2410-16033@token-level-sampling1,DBLP:journals/corr/abs-2410-11744@token-level-sampling2} within single-paradigm reasoning paths, we purpose sequential paradigm-level sampling method, named SMPS. 
This allows the model to generate outputs by sampling across different paradigms.
For instance, in a two-paradigm reasoning scenario, the model first instantiates $J$ distinct paths for the initial reasoning paradigm $\tau_1$.
\begin{equation}
\tau_{1j} \sim \mathbb{P}(\tau_{1j} \mid x), \quad \forall j \in {1, \dots, J}
\end{equation}

Subsequently, for each of these $\tau_1$ paths, the model instantiates $K$ possible paths for the secondary reasoning paradigm $\tau_2$.
\begin{equation}
\tau_{2k} \sim \mathbb{P}(\tau_{2k} \mid x, \tau_{1j}), \quad \forall k \in {1, \dots, K}, \forall j
\end{equation}

This hierarchical sampling process yields a total of $J\times K$ potential responses, denoted as $y$.
\begin{equation}
y_{jk} \sim \mathbb{P}(y_{jk} \mid x, \tau_{1j}, \tau_{2k}), \quad \forall j, k
\end{equation}

Overall, the SMPS method utilizes a combinatorial expansion of reasoning paths to explore a diverse paradigm-based solution space, enhancing the robustness and depth of the reasoning results.

\section{Experimental Settings}

\subsection{Evaluation Setup}

\nbf{Datasets.}
We conduct extensive experiments to evaluate the model's mathematical reasoning abilities, including arithmetic computation and theorem proving.
The arithmetic computation ability is evaluated on the GSM8K~\citep{DBLP:journals/corr/abs-2110-14168@gsm8k}, MATH~\citep{DBLP:conf/nips/HendrycksBKABTS21@math}, AMC2023~\citep{amc2023} and AIME2024~\citep{aime2024} datasets. 
Theorem proving ability is evaluated on the miniF2F test set~\citep{DBLP:conf/iclr/ZhengHP22@minif2f}, which features mathematical problems of Olympiad difficulty.
Further details are in~\cref{ass:benchmarks}.

\begin{table*}[!tp]
\small
\centering
\begin{adjustbox}{max width=0.86\textwidth}
\begin{tabular}{lccccc|cc}
\toprule
\multirow{3}{*}{\centering Model} & \multirow{3}{*}{\centering Model Types} & \multicolumn{4}{c|}{\textbf{Arithmetic Computation}} & \multicolumn{2}{c}{\textbf{Theorem Proving}} \\
\cmidrule(lr){3-6} \cmidrule(lr){7-8}
& & MATH & GSM8k & AMC2023 & AIME2024 & \multirow{2}{*}{\centering \begin{tabular}[c]{@{}c@{}}miniF2F-test\\ Sample Budget ($N$)\end{tabular}} & miniF2F-test \\
& & Pass@1 & Pass@1 & Maj@64 & Maj@64 & & Pass@$N$ \\  
                                \midrule
\rowcolor{RowGray} o1-mini                          & Proprietary                                                                                                & 90.0{$^{\emptyset}$}   & 94.8{$^{\emptyset}$}   & 38/40{$^{\emptyset}$}  & 17/30{$^{\emptyset}$}  & 1                                                                                                   & 13.2     \\
\rowcolor{RowGray} GPT-4                           & Proprietary                                                                                             & 42.5{$^{\emptyset}$}   & 87.1{$^{\emptyset}$}   & 25/40{$^{\emptyset}$}  & 6/30{$^{\emptyset}$}  & 128                                                                                                   & 24.6    \\
\rowcolor{RowGray} GPT-4o                          & Proprietary                                                                                                & 76.6{$^{\emptyset}$}   & 90.5{$^{\emptyset}$}   & 24/40{$^{\emptyset}$}  & 3/30{$^{\emptyset}$} & 128                                                                                                   & 25.0   \\
\midrule
Llama-3.1-8B                    & Foundation                                                                                               & 4.2{$^{\emptyset}$}    & 6.2{$^{\emptyset}$}    & 1/40{$^{\emptyset}$}  & 0/30{$^{\emptyset}$}  & 128                                                                                                   & 25.8    \\
Llama-3.1-8B-Instruct           & Foundation                                                                                                & 47.2{$^{\emptyset}$}   & 76.6{$^{\emptyset}$}   & 16/40{$^{\emptyset}$}  & 5/30{$^{\emptyset}$}  & 128                                                                                                   & 23.4    \\
Mistral-7B                      & Foundation                                                                                                  & 14.3   & 40.3  & 5/40{$^{\emptyset}$}  & 0/30{$^{\emptyset}$}   & 1$\times$32$\times$100                                                                                                   & 22.1    \\
Mixtral-8x7B                    & Foundation                                                                                                & 28.4   & 74.4   & 8/40{$^{\emptyset}$}  & 0/30{$^{\emptyset}$} & 1$\times$32$\times$100                                                                                                    & 23.4    \\
\midrule
MUSTARD                         & GMM                                                                                               & 13.8{$^{\emptyset}$}   & 27.9{$^{\emptyset}$}   & -  & - & 1                                                                                                   & 7.8     \\
Llemma-7B                       & GMM                                                                                               & 18.6   & 41.0          & 2/40{$^{\emptyset}$}  & 0/30{$^{\emptyset}$} & 1$\times$32$\times$100                                                                                                    & 26.2    \\
DeepSeekMath-7B-Base    & GMM                                                                                              & 11.8{$^{\emptyset}$}   & 22.2{$^{\emptyset}$}   & 3/40{$^{\emptyset}$}  & 0/30{$^{\emptyset}$} & 1$\times$32$\times$100                                                                                                    & 28.3    \\
InternLM2-Math-7B-Base          & GMM                                                                                             & 21.5   & 49.2   & 6/40{$^{\emptyset}$}  & 0/30{$^{\emptyset}$} & 1$\times$32$\times$100                                                                                                    & 30.3    \\
InternLM2-Math-Plus-7B          & GMM                                                                                              & 53.0{$^{\emptyset}$}   & 85.8{$^{\emptyset}$}   & 15/40{$^{\emptyset}$}  & 1/30{$^{\emptyset}$} & 1$\times$32$\times$100                                                                                                    & 43.3    \\ 
\midrule
\multirow{3}{*}{\modelname}           &  \multirow{3}{*}{GMM}                                                                                                 & \multirow{3}{*}{\textbf{66.7}{$^{\emptyset}$}}   & \multirow{3}{*}{\textbf{88.7}{$^{\emptyset}$}}   & \multirow{3}{*}{\textbf{34/40}{$^{\emptyset}$}}  & \multirow{3}{*}{\textbf{12/30}{$^{\emptyset}$}}  & 128$\times$1                                                                                          & 52.9{$^{\emptyset}$}   \\
 & & & & & & 32$\times$100 & 59.4{$^{\emptyset}$} \\
& & & & & & 128$\times$128 & \textbf{66.0}{$^{\emptyset}$} \\
\bottomrule
\end{tabular}
\end{adjustbox}
\caption{A overall comparison of \modelname with five types of general mathematical reasoners (Proprietary, Foundational, and General-purpose Mathematical Models (GMM)) on three mathematical benchmarks. 
Results are shown for zero-shot (denoted by $\emptyset$) or few-shot settings by default. 
For the miniF2F benchmark, we report the best results from the relevant literature with a specified sample budget. 
\textbf{Bolded} scores are the best performance among all models except for the proprietary one.}
\label{table:general-purpose-results}
\end{table*}

\nbf{Metrics.}
Accuracy is the primary evaluation metric.
For arithmetic tasks, we adopt widely-used CoT settings~\citep{DBLP:conf/nips/Wei0SBIXCLZ22@cot}, rounding answers to the nearest integer. 
The SymPy library\footnote{https://www.sympy.org/} is used for parsing and evaluation. 
To handle numerical representation variations, the model explicitly states the final answer~\citep{numina_math_dataset@numina}. 
For theorem proving, we follow the recent advances~\citep{DBLP:journals/corr/abs-2408-08152@deepseek-prover-v1.5}, adapting the miniF2F benchmark from Lean 3 to Lean 4.
The pass@$N$ metric evaluates proof correctness within $N$ attempts. 
Our \modelname uses SMPS with NLR and SR, where $N = N_{\text{NLR}} \times N_{\text{SR}}$. 
Model settings are detailed in~\cref{ass:details_metrics}.

\subsection{Implementation Details}

We fine-tuned widely-used DeepSeekMath-Base 7B~\citep{DBLP:journals/corr/abs-2402-03300@deepseekmath} and Llama-3.1 8B~\citep{DBLP:journals/corr/abs-2407-21783@llama3} models, employing our PPT method on \datasetname dataset.
Unless otherwise specified, \modelname model is based on DeepSeekMath-Base 7B.
The details are available in~\cref{ass:training-settings}.

\subsection{Baselines}

We examine three categories of baseline models, with results reported using CoT prompting.

\nbf{General-purpose mathematical models.}
To evaluate \methodname's  generalization, we includes Mustard~\citep{DBLP:conf/iclr/HuangLLCXWLSL24@mustard}, DeepSeekMath~\citep{DBLP:journals/corr/abs-2402-03300@deepseekmath}, InternLM-Math~\citep{ying2024internlmmath@internlm-math}, Llama-3.1~\citep{DBLP:journals/corr/abs-2407-21783@llama3}, Mistral~\citep{DBLP:journals/corr/abs-2310-06825@mistral-7b}, and Llemma~\citep{DBLP:conf/iclr/AzerbayevSPSMJD24@llemma}.

\nbf{Task-specific mathematical models.}
We consider several expert models on mathematical optimization for specific tasks, such as large-scale mathematical data and inference search improvements.
The arithmetic experts encompass Qwen2.5-Math~\citep{DBLP:journals/corr/abs-2409-12122@qwen25-math}, WizardMath~\citep{DBLP:journals/corr/abs-2308-09583@wizardmath}, MetaMath~\citep{DBLP:conf/iclr/YuJSYLZKLWL24@metamath}, DART-Math~\citep{DBLP:journals/corr/abs-2407-13690@dart}, InternLM-Math~\citep{ying2024internlmmath@internlm-math},
, DeepSeekMath-Instruct / RL~\citep{DBLP:journals/corr/abs-2402-03300@deepseekmath}, Xwin-Math~\citep{DBLP:journals/corr/abs-2403-04706@xwin-math}, ToRA~\citep{DBLP:conf/iclr/GouSGSYHDC24@tora}, and NuminaMath~\citep{numina_math_dataset@numina}.
The theorem proving experts include LLM-Step~\citep{DBLP:journals/corr/abs-2310-18457@llmstep}, GPT-f~\citep{DBLP:journals/corr/abs-2009-03393@gpt-f}, Lean-STaR~\citep{DBLP:journals/corr/abs-2407-10040@lean-star}, Hypertree Proof Search~\citep{DBLP:conf/nips/LampleLLRHLEM22@hypertree}, DeepSeek-Prover~\citep{DBLP:journals/corr/abs-2408-08152@deepseek-prover-v1.5}, and InternLM2.5-StepProver~\citep{DBLP:journals/corr/abs-2410-15700@internlm25_step}.

\nbf{Foundation and proprietary models.}
We present open-source foundation models Llama-3.1~\citep{DBLP:journals/corr/abs-2407-21783@llama3}, Mistral~\citep{DBLP:journals/corr/abs-2310-06825@mistral-7b}, and Mixtral~\citep{DBLP:journals/corr/abs-2401-04088@mixtral}, along with proprietary models o1-mini~\citep{openai2024openaio1card@o1}, GPT-4~\citep{DBLP:journals/corr/abs-2303-08774@gpt4}, and GPT-4o~\citep{DBLP:journals/corr/abs-2410-21276@gpt4o}.

\section{Main Results}
\label{sec:main_result}

\subsection{Comparisons with General-purpose Mathematical Models}

To evaluate the comprehensive mathematical reasoning abilities of \modelname, we compare it with widely-used models and SOTA mathematical models, which are based on single reasoning paradigms. 
As shown in ~\cref{table:general-purpose-results}, \modelname outperforms others across five challenging benchmarks in a zero-shot setting, highlighting its strong cross-task versatility. 
The main findings are:
(1) arithmetic computation: \modelname achieves the best results, outperforming InterLM2-Math-Plus-7B by an absolute $13.7\%$ on MATH and $2.9\%$ on GSM8K.
Compared to Llama-3.1-8B-Instruct, it achieves an increase in correct answers on AMC2023 ($34$ vs $16$) and AIME2024 ($12$ vs $5$).
(2) theorem proving: \modelname sets a new zero-shot benchmark on miniF2F, even surpassing GPT-4o model by a $41.0\%$ absolute increase in few-shot setting. 
(3) zero-shot vs few-shot: \modelname's zero-shot performance exceeds all few-shot results from other models, with an absolute $37.7\%$ improvement over DeepSeekMath-7B-Base on miniF2F.
These findings indicate the generalization capability and comprehensive mathematical reasoning ability of the \methodname framework.
This suggests NLR descriptions and SR verification rehearse precise mathematical reasoning for AR. 
Additionally, we provide qualitative analysis of error cases in ~\cref{ass:error-case}.
In summary, \methodname effectively handles diverse reasoning challenges, with synergies across paradigms boosting its overall performance.

\begin{table}[!tp]
\small
\centering
\vspace{1em}
\begin{adjustbox}{max width=0.48\textwidth}
\begin{tabular}{l|c|c}
\toprule
Model & Sample Budget $N$ & miniF2F \\ \midrule
LLMStep & 1$\times$32$\times$100 & 27.9 \\
GPT-f & 64$\times$8$\times$512 & 36.6 \\
Hypertree Proof Search & 64$\times$5000 & 41.0 \\
Lean-STaR & 64$\times$1$\times$50 & 46.3 \\
DeepSeek-Prover-V1.5-Base & 6400 & 42.2 \\
DeepSeek-Prover-V1.5-SFT + RMaxTS & 4$\times$6400 & 56.3 \\
DeepSeek-Prover-V1.5-SFT + RMaxTS & 32$\times$6400 & 60.2 \\
DeepSeek-Prover-V1.5-RL + RMaxTS & 4$\times$6400 & 59.6 \\
DeepSeek-Prover-V1.5-RL + RMaxTS & 32$\times$6400 & 63.5 \\
InternLM2.5-StepProver & 64$\times$32$\times$100 & 54.5 \\
InternLM2.5-StepProver-BF & 1$\times$32$\times$600 & 47.3 \\
InternLM2.5-StepProver-BF & 256$\times$32$\times$600 & 59.4 \\
InternLM2.5-StepProver-BF+CG & 2$\times$32$\times$600 & 50.7 \\
InternLM2.5-StepProver-BF+CG & 256$\times$32$\times$600 & 65.9 \\
\midrule
\modelname & 128$\times$128 & \textbf{66.0}{$^{\emptyset}$} \\
\bottomrule
\end{tabular}
\end{adjustbox}
\caption{The zero-shot ($\emptyset$) performance of \modelname and the few-shot performance of theorem proving optimized models on the miniF2F benchmark. 
The highest scores are in \textbf{bold}.}
\label{table:proving-res}
\vspace{-2em}
\end{table}

\subsection{Comparisons with Theorem Proving Experts}

\cref{table:proving-res} shows \modelname's impressive zero-shot performance on miniF2F, achieving $66.0\%$ accuracy without demonstrations, while competitors require few-shot examples.
Its zero-shot performance is noteworthy when considering computational efficiency. 
For instance, under similar computational constraints, \modelname surpasses DeepSeek-Prover-V1.5-RL + RMaxTS by $6.4\%$ absolute points ($66.0\%$ vs $59.6\%$ with $128\times128$ vs $4\times6400$) and InternLM2.5-StepProver-BF+CG by $15.3\%$ absolute points ($66.0\%$ vs $50.7\%$ with $128\times128$ vs $2\times32\times600$).
Even with larger sample sizes, \modelname remains superior, surpassing few-shot InternLM2.5-StepProver-BF+CG by $0.1\%$, despite considering $300$ times more solutions. 
These results highlight the efficiency of multi-paradigm reasoning and \methodname's ability to explore paradigm-based solution spaces.

\vspace{-1ex}
\begin{table}[tp]
\small
\centering
\begin{adjustbox}{max width=0.88\textwidth}
\begin{tabular}{lcc}
\toprule
\multirow{2.5}{*}{Model} & MATH & GSM8k  \\
\cmidrule{2-3}
 & ZS Pass@1 & ZS Pass@1\\
\midrule
WizardMath & 10.7 & 54.9  \\
MetaMath-7B  & 19.8 & 66.5 \\
MetaMath-Llemma-7B & 30.0 & 69.2 \\
MetaMath-Mistral-7B & 28.2 & 77.7\\
ToRA & 40.1 & 68.8  \\
ToRA-CODE & 44.6 & 72.6 \\
NuminaMath-7B-CoT & $54.4^{\dagger}$ & $66.6^{\dagger}$ \\
Xwin-Math-7B & 40.6 & 82.6 \\
DeepSeekMath-Instruct-7B  & 46.8 & 73.6 \\
NuminaMath-7B-TIR & $55.3^{\dagger}$ & $73.6^{\dagger}$ \\
DeepSeekMath-RL-7B & 51.7 & 88.2\\
DART-Math-7B & 53.6 & 86.6 \\
Qwen2.5-Math-7B-Instruct & \textbf{83.6} & \textbf{95.2} \\
\midrule
\modelname & {\ul 66.7} & {\ul 88.7} \\
\bottomrule
\end{tabular}
\end{adjustbox}
\caption{Zero-shot (ZS) performance on the MATH and GSM8K benchmarks for the arithmetic task. $\dagger$ denotes our reported results with the open-sourced model weights. The best results are in \textbf{bold}, and the second-best results are {\ul underlined}.}
\label{table:arithmetic-res}
\vspace{-0.5em}
\end{table}

\subsection{Comparisons with Arithmetic Experts}
\label{ss:comparison-ac}

Since several mathematical reasoners are primarily trained on arithmetic computation rather than formal theorem proving, and often struggle with the latter, we classify them as experts in arithmetic computation to enable a fair comparison. 
\cref{table:arithmetic-res} highlights \modelname's strong performance in arithmetic computation, surpassing expert models and achieving a $15\%$ improvement over RL-based methods, with DeepSeekMath-RL-7B as the representative, on the MATH benchmark for arithmetic tasks.
Unlike methods such as ToRA and NuminaMath, which rely on code without complete reasoning, \modelname leverages multi-paradigm reasoning to outperform them across all benchmarks.
For example, on the MATH dataset, \modelname achieves an $11.4\%$ absolute improvement over the tool-integrated NuminaMath-7B-TIR, underscoring the effectiveness of complete code-based reasoning over fragmented interleaving of natural language and code.
To further explore the relationship between resource utilization (training sample size) and performance, we fit a quadratic function to the current SOTA models and perform a Pareto frontier analysis~\citep{Lotov2008@pareto}, as shown in~\cref{fig:intro-result} (Details in~\cref{ass:details-arithmetic}).
\modelname outperforms the optimal performance of single-paradigm methods with similar data volume, suggesting a new optimal curve for multi-paradigm reasoning that surpasses single-paradigm approaches.

\section{Ablation Study}

\begin{figure}
\centering
\includegraphics[width=0.48\textwidth]{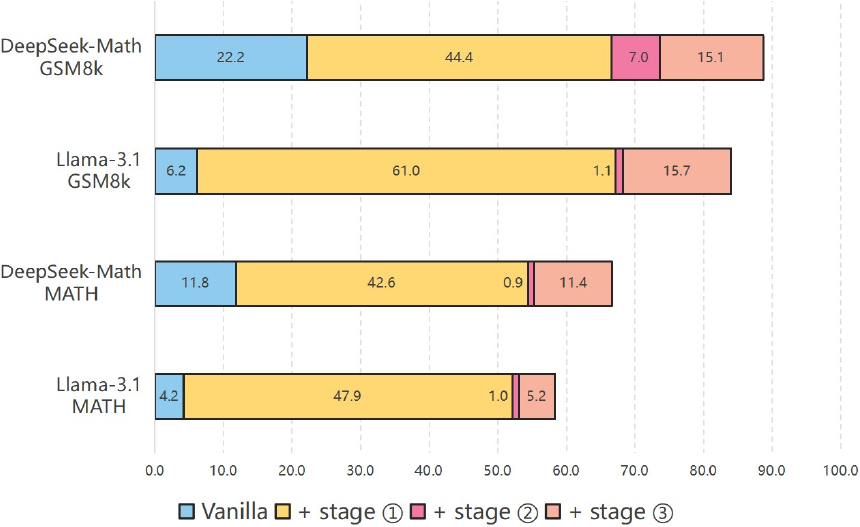}
\caption{An evaluation of the effectiveness of the PPT strategy.
We present the zero-shot Pass@1 results on the MATH and GSM8k benchmarks across three cumulative stages of the PPT strategy.
The results highlight the PPT strategy's cumulative effectiveness, showing increased performance with each progressive stage.
}
\label{fig:ppt-res}
\vspace{-1em}
\end{figure}

\subsection{Impact of Stages in PPT Method}
\label{ss:ppt_stages}

We evaluate the PPT method using two base models, Llama-3.1-8B and DeepseekMath-7B-Base, on the MATH and GSM8K benchmarks at different PPT stages.
As shown in~\cref{fig:ppt-res}, the vanilla models exhibit limited performance.
After stage \ding{172}, performance improves significantly, with Llama-3.1-Base gaining $47.9\%$ on MATH and $61.0\%$ on GSM8K, emphasizing the need for mathematical understanding and model warming.
Stage \ding{173}, utilizing NLR and AR data, yields limited gains compared to stage \ding{172} due to pre-training data already containing relevant content. 
However, this does not imply that stage \ding{173} is unimportant, as it plays a critical role in CoR by reactivating and refining computational skills essential for collaborating with other reasoning paradigms.
Both stages \ding{172} and \ding{173} focus on reactivating models for mathematical tasks.
In stage \ding{174}, training with three paradigms further enhances performance, suggesting that rare or unseen paradigms improve reasoning abilities.
The consistent performance gains with added paradigms in ~\cref{fig:ppt-res} robustly demonstrate the superiority of our multi-paradigm PPT approach over single- or fewer-paradigm fine-tuning of the same base model, yielding a more powerful, comprehensive mathematical reasoner.

\begin{table}[tp]
\small
\centering
\begin{tabular}{l|c|c}
\toprule
\multirow{2}{*}{Benchmark} & \multirow{2}{*}{\makecell{NLR$\rightarrow$ \textbf{AR} $\rightarrow$ \\\textbf{SR} $\rightarrow$ Summary}} & \multirow{2}{*}{\makecell{NLR$\rightarrow$ \textbf{SR} $\rightarrow$ \\\textbf{AR} $\rightarrow$ Summary}} \\ 
&  & \\ \midrule
MATH  & 49.9 & \textbf{66.7} \\
GSM8K & 84.2 & \textbf{88.7} \\
\bottomrule
\end{tabular}
\caption{Zero-shot Pass@1 results for varying paradigm orders on the MATH and GSM8k benchmarks. The best results are in \textbf{bold}.}
\label{table:order-res}
\vspace{-0.5em}
\end{table}

\begin{table*}[!tp]
\small
\centering
\begin{adjustbox}{max width=0.8\textwidth}
\begin{tabular}{l|cccc|c}
\toprule
Model & GSM8k@1 & MATH@1 & AMC2023 (Maj@64) & AIME2024 (Maj@64) & miniF2F@128 \\ \midrule
DSM + NLR        & 33.9    & 27.8   & 14/40                                                       & 1/30                                                         & -           \\
DSM + AR         & 75.8    & 37.6   & 10/40                                                       & 0/30                                                         & -           \\
DSM + SR         & -       & -      & -                                                           & -                                                            & 44.3        \\
DSM + CoR (ours) & 88.7    & 66.7   & 34/40                                                       & 12/30                                                        & 52.9        \\ \bottomrule
\end{tabular}
\end{adjustbox}
\caption{Zero-shot results of multi- and single-paradigm fine-tuning on DeepSeekMath-7B-Base (DSM) for mathematical Reasoning benchmarks. The best results are in \textbf{bold}.}
\label{table:deepseek-comparison}
\end{table*}

\begin{table}[tp]
\small
\centering
\vspace{-0.5em}
\setlength{\tabcolsep}{4pt}
\begin{tabular}{lc|cc|c}
\toprule
Model & Size & MATH & GSM8k & miniF2F \\
\midrule
\multirow{2}{*}{Qwen2.5-Math (Base)} & 1.5B & 34.0 & 39.3 & 0.0 \\
& 7B & 51.8 & \textbf{90.0} & 0.0 \\
\multirow{2}{*}{Qwen2.5-Math (CoR)} & 1.5B & 57.6 & 84.5 & 51.6 \\
& 7B & 64.7 & \textbf{90.0} & 52.5 \\
\midrule
\multirow{2}{*}{Llama-3.1 (Base)} & 8B & 4.2 & 6.2  & 25.8 \\
& 70B & 16.8 & 20.5 & 22.5 \\
\multirow{2}{*}{Llama-3.1 (CoR)} & 8B & 58.2 & 84.0  & 53.3 \\
& 70B & \textbf{70.7}  & \textbf{90.0} & \textbf{56.2} \\
\bottomrule
\end{tabular}
\caption{Zero-shot Pass@1 results on the mathematical benchmarks across different model scales. The best scores are in \textbf{bold}.}
\label{table:scales-res}
\vspace{-1.0em}
\end{table}

\subsection{Order of the Reasoning Paradigms}

As shown in~\cref{table:order-res}, we investigate the impact of varying the sequence of reasoning paradigms in \modelname during zero-shot arithmetic tasks, modifying only the prompt structure. 
Given that NLR closely aligns with the pre-training of LLMs, it serves as the first paradigm.
The results indicate that SR before AR resulted in the highest accuracy, with $66.7\%$ on MATH, compared to $49.9\%$ for AR before SR. 
A possible explanation is that SR plays a crucial role in decomposing the problem into manageable sub-steps, thereby providing a structured foundation for subsequent AR. 
This demonstrates that paradigms in CoR are interdependent, not isolated, and performance hinges on how they are linked during reasoning, beyond merely mastering individual paradigms.
Furthermore, positioning AR preceding the Summary boxes may facilitate a more coherent integration of the computational outcome into the final answer. 

\subsection{Impact of Model Scales}

\cref{table:scales-res} evaluates base model performance under the \methodname framework with varying parameters, such as Qwen2.5-Math models ($1.5$B / $7$B) and Llama-3.1-8B ($8$B / $70$B).
The results show that \methodname scales with model size. 
The $7$B Qwen2.5-Math outperforms the $1.5$B version by $7.1\%$ on MATH and $5.5\%$ on GSM8K. 
Similarly, Llama-3.1 $70$B gains $2.9\%$ over its $8$B version on minF2F.
This supports that \methodname exhibits parameter scalability.

\subsection{Multi- vs. Single-Paradigm Reasoning}

To further validate the superior performance of multi-paradigm reasoning, we conducted an ablation experiment comparing it to single-paradigm fine-tuning
We fine-tuned the DeepSeekMath-7B-Base model separately using single-paradigm paths (NLR, AR, SR) extracted from the MPM dataset. 
All experiments employed identical base models, problem sets, and training data sizes ($10,000$ samples per paradigm). 
Experimental results, summarized in Table~\ref{table:deepseek-comparison}, clearly demonstrate that our CoR model significantly outperforms single-paradigm fine-tuned models across diverse tasks. 
Specifically, for GSM8k and MATH datasets, CoR achieves improvements of $12.9\%$ and $29.1\%$, respectively, compared to the best-performing single-paradigm approaches. 
On AMC2023 and AIME2024, CoR solves $20$ and $11$ more problems, respectively, compared to single-paradigm approaches. 
Lastly, on minF2F, CoR achieves an accuracy improvement of $8.6\%$. 
Collectively, these improvements underline CoR's effectiveness in integrating multiple reasoning paradigms, enabling robust and generalized problem-solving capabilities.

\section{Conclusions}

This paper introduces CoR, a novel unified reasoning framework that enhances mathematical reasoning in LLMs by synergistically integrating natural language, algorithmic, and symbolic. 
CoR tackles the limitations of single-paradigm approaches with the PPT strategy and the MPM dataset, enabling LLMs to progressively master diverse reasoning paradigms for improved generalization and performance.
Our \modelname outperformed SOTA models on five challenging benchmarks, showcasing enhanced zero-shot generalization. 
Ablation studies validated the benefits of progressive training and paradigm sequence.
Additionally, CoR offers a new perspective on test-time scaling through multi-paradigm reasoning to enhance both efficiency and performance.

\section*{Acknowledgments}

This work was partly supported by the Shenzhen Science and Technology Program ( JSGG20220831110203007 ) and  the  research grant No. CT20240905126002 of the Doubao Large Model Fund.

\section*{Limitations}

This paper proposes a general zero-shot reasoning method that supports different paradigms to solve multiple tasks. 
Evaluating zero-shot performance is challenging, as many methods rely on a single evaluation metric, like zero-shot pass@1. 
Additionally, most approaches focus on few-shot settings, as seen in benchmarks like miniF2F, limiting the comprehensiveness of our evaluation with aligned settings. 
To address this, we include additional experiments in~\cref{ass:diff-eval-strategies} to examine the impact of different evaluation strategies. 
On the other term, \methodname can explore a multi-paradigm solution space, covering up to three paradigms, such as in large-scale arithmetic tasks. 
However, due to resource constraints and the fact that other methods can only predict within a single paradigm, we did not further consider using SMPS on benchmarks such as MATH to ensure fairness.

\bibliography{custom}

\clearpage
\appendix
\section*{Appendix}

\section{Discussion on Reasoning Hierarchy: Paradigms, Paths, and Steps}
\label{ass:reasoning_hierarchy}

This section attempts to explore the essence of reasoning, contrasts CoR with current methods, examines the factors contributing to its effectiveness, and provides a new perspective for future research.

\begin{figure}[!tp]
\centering
\includegraphics[width=0.48\textwidth]{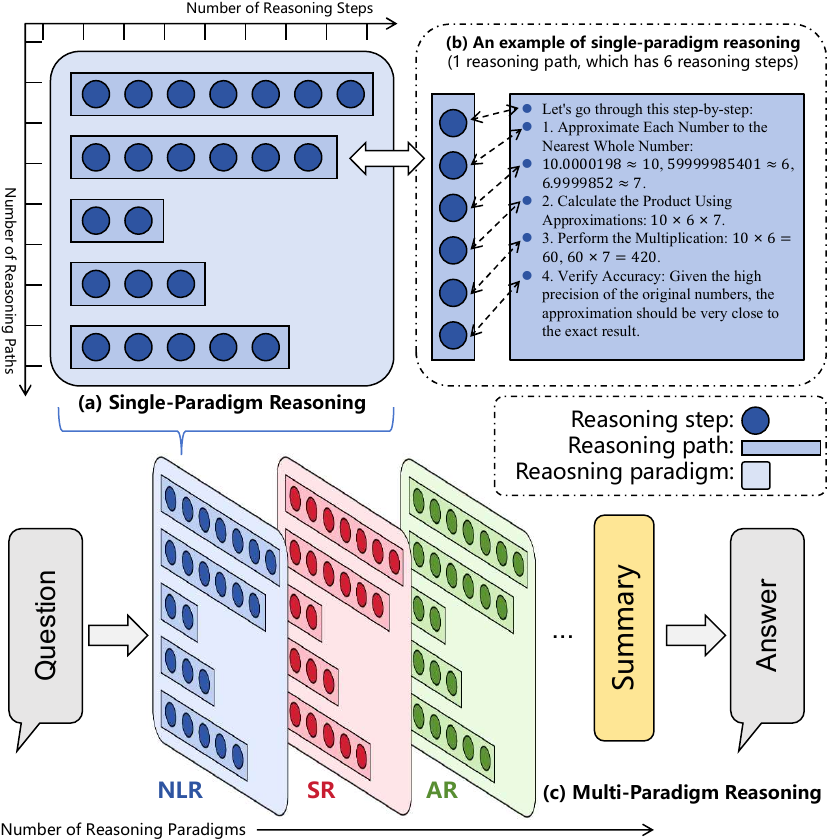}
\caption{An overview of the reasoning hierarchical structure. (a) A single reasoning paradigm depicts multiple distinct reasoning paths. (b) An example of single-paradigm reasoning includes one reasoning path, which contains some reasoning steps. (c) Multi-paradigm reasoning includes several distinct reasoning paradigms.}
\label{fig:reasoning-hier}
\end{figure}

As shown in~\cref{fig:reasoning-hier}, this paper posits that reasoning texts generated by LLMs exhibit a reasoning hierarchical structure, which consists of three levels: reasoning paradigms, reasoning paths, and reasoning steps. 
\begin{itemize}
\setlength{\itemsep}{1pt}
\setlength{\parskip}{2pt}
\setlength{\parsep}{0pt}
\item \textbf{Reasoning steps} represent the fundamental units, each comprising one or more tokens and encompassing an incomplete stage of the solution process. 
\item \textbf{Reasoning paths} consist of several reasoning steps, forming a complete line of reasoning that typically includes a final answer and the solution process. 
\item \textbf{Reasoning paradigms} comprise one or more reasoning paths. They often contain multiple potential final answers, thus necessitating a summarization method, such as a summary module, to extract the ultimate answer. Furthermore, a reasoning paradigm uses a single knowledge media, such as natural language. 
\end{itemize}

\begin{figure}[!t]
\centering
\includegraphics[width=0.48\textwidth]{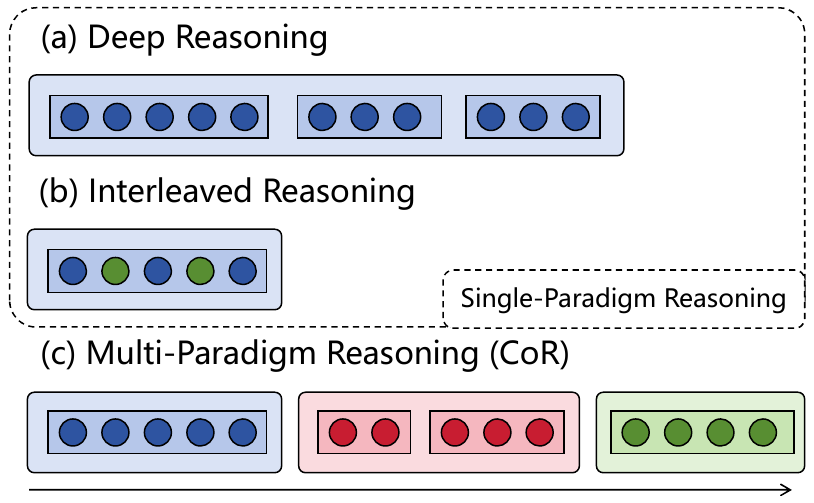}
\caption{A comparison of reasoning methods in several advanced studies.}
\label{fig:reasoning-examples}
\end{figure}

Furthermore, current studies can be categorized based on their focus. 
As shown in~\cref{fig:reasoning-hier} (a), contemporary work concentrates within a single paradigm, optimizing along two dimensions: depth (the number of reasoning steps) and width (the number of reasoning paths). 
For instance, regarding reasoning depth, the CoT method~\citep{DBLP:conf/nips/Wei0SBIXCLZ22@cot} employs prompts to increase intermediate steps within one reasoning path to achieve higher performance. 
Concerning reasoning width, some approaches~\citep{DBLP:conf/iclr/0002WSLCNCZ23@self-consistency} involve altering sampling techniques to generate multiple distinct reasoning paths.
Random sampling exemplifies this. 
Other studies~\citep{DBLP:conf/acl/ZhuWZZ0GZY23@core} employ scoring mechanisms for reasoning steps, guiding LLMs to generate several more desirable reasoning paths. 
Monte Carlo search illustrates this. 
Building upon these generated reasoning paths, existing work proposes various integration strategies to obtain the final answer, such as Best-of-N and Self-consistency~\citep{DBLP:conf/iclr/0002WSLCNCZ23@self-consistency}. 
Specifically, as shown in~\cref{fig:reasoning-examples} (a), recent advanced studies introduce the deep reasoning method, which focuses on generating serial concatenation of reasoning paths followed by summarization (like OpenAI o1~\cite{openai2024openaio1card@o1}). 
Since these methods are based on different optimization dimensions, we can combine them feasibly in applications.

Specifically, some approaches prioritize a dominant paradigm for reasoning yet hope to integrate other paradigms for guidance. 
For instance, as shown in~\cref{fig:reasoning-examples} (b), the reasoning process of ToRA~\citep{DBLP:conf/iclr/GouSGSYHDC24@tora} presents an interleaved reasoning method with different knowledge media.
It involves initial natural language generation, which is followed by code generation. 
Subsequently, the process awaits the results of code execution before generating further natural language. 
InternLM2.5-StepProver~\citep{DBLP:journals/corr/abs-2410-15700@internlm25_step} exhibits a similar pattern. 
It incorporates natural language annotations to support symbolic reasoning steps. 
However, these interleaved reasoning approaches do not constitute true multi-paradigm reasoning. 
The primary paradigm can independently achieve the final answer without the supplementary paradigms. 
Consequently, these methods imply a single-paradigm reasoning approach enhanced by other paradigms, rather than genuine multi-paradigm reasoning.

Current methods often \textit{overlook the synergistic effects between paradigms and underestimate the importance of complete reasoning for achieving accurate results}. 
To address this challenge, as shown in~\cref{fig:reasoning-hier} (c) and~\cref{fig:reasoning-examples} (c), we introduce Chain-of-Reasoning (CoR), a unified reasoning framework capable of multi-paradigm reasoning. 
This framework embodies the following potential advantages:

\nbf{Sequential reasoning dependency}. 
As a type of multi-paradigm reasoning, CoR is not a mere accumulation of isolated steps; instead, it represents a coherent, interconnected process. 
The output from an earlier paradigm functions not only as input for a subsequent paradigm but also informs the foundational information for its reasoning. 
For instance, knowledge derived from a natural language paradigm guides subsequent algorithmic paradigms, thereby enhancing the efficiency and accuracy of code generation based on detailed natural language reasoning. 
This mechanism also provides substantial context from prior paradigms to later paradigms. 
It operates as a form of preliminary rehearsal where subsequent paradigms can follow correct reasoning steps or rectify incorrect ones.

\nbf{A novel direction for test-time scaling emerges}. 
We observe that the achievable improvements of single-paradigm reasoning are increasingly constrained by various search methods applied to the solution space.
For instance, DeepSeek-Prover-V1.5-RL + RMaxTS requires $32\times 6400$ reasoning paths to achieve $63.5\%$ few-shot accuracy on the miniF2F benchmark. 
While extending the number of candidate solution paths seems a natural way to enhance the hit rate for ground truth solutions, the substantial computational effort involved underscores the inherent limitations of single-paradigm reasoning.
This observation suggests that scaling up reasoning within a single paradigm during test time is inadequate for addressing complex tasks, such as mathematical problem solving. 
From a broader perspective, our CoR framework reexamines this challenge by shifting the focus of scaling efforts from reasoning paths to reasoning paradigms, thereby proposing a novel avenue for advancement.

\nbf{Expanding the solution space}.
In multi-paradigm reasoning, the solution space is expanded by considering potential orthogonal relationships between reasoning paradigms. 
This allows generated solutions to be searched within different paradigms. 
Moreover, diverse solutions across paradigms can leverage chain relationships in reasoning to mutually inform each other. 
Therefore, our CoR framework enhances both intra-paradigm and inter-paradigm search capabilities within a large solution space. 
This diversity increases the likelihood of discovering optimal solutions.

\nbf{Journey-based reasoning learning}.
CoR enhances models' learning trajectory, allowing them to perform deep reasoning both within specific paradigms and across multiple paradigms. 
This approach expands the ``journey'' by introducing the ability to navigate and collaborate across diverse reasoning paradigms.

\nbf{Compatibility with existing methods}.
Since most current approaches are fundamentally based on a single paradigm, CoR can seamlessly integrate these methods within its specified paradigms, enabling them to work synergistically and leverage their strengths. 
For example, CoT can be incorporated as a specific implementation of the NLR paradigm in CoR. 
This implies that CoR serves as a platform that facilitates continuous integration and collaboration.

\section{Experiment Details}

In this study, we use open-source tools, models, and datasets in compliance with their respective open-source licenses and solely for academic research purposes.

\subsection{The detail of the Universal Text Template}
\label{ass:universal-template}

\begin{figure*}[tp]
\begin{tcolorbox}[
    colback=white,
    colframe=black!50,
    boxrule=0.5mm,
    arc=0.5mm,
    outer arc=0.5mm,
    title=Universal Template Designed for Multi-paradigm Reasoning.
]
\begin{tcolorbox}[
    boxrule=0.5mm,
    arc=0.5mm,
    outer arc=0.5mm,
    colback=problem-title,
    colframe=problem-title!75,
    title=\#\#\# Problem:
]
[Statement of the problem]
\end{tcolorbox}

\begin{tcolorbox}[
    boxrule=0.5mm,
    arc=0.5mm,
    outer arc=0.5mm,
    colback=nlr-back,
    colframe=nlr-title,
    title=Let's go through this step-by-step:
]
\begin{itemize}
    \item[1.] [Step 1 description]
    \item[2.] [Step 2 description]
    \item[3.] [Step 3 description]
    \item[$\cdots$] [Further steps as needed]
    \item[$\checkmark$] [Verification or conclusion of the step-by-step reasoning]
\end{itemize}
\end{tcolorbox}

\begin{tcolorbox}[
    boxrule=0.5mm,
    arc=0.5mm,
    outer arc=0.5mm,
    colback=sr-back!75,
    colframe=sr-title,
    title=Let's write the corresponding formal proof in Lean 4 to prove this:\\\#\#\# Formal proof in Lean 4
]
\begin{verbatim}
[Lean 4 code for formal proof]
\end{verbatim}
\hfill
\begin{tikzpicture}
    \node[fill=white,draw] (lean) {\textcolor{gray}{[Lean 4 Output]}};
\end{tikzpicture}
\end{tcolorbox}

\begin{tcolorbox}[
    boxrule=0.5mm,
    arc=0.5mm,
    outer arc=0.5mm,
    colback=ar-title,
    colframe=ar-back,
    title=Let's use Python to perform these calculations:\\\#\#\# Code in Python
]
\begin{verbatim}
[Python code for calculation]
\end{verbatim}
\hfill
\begin{tikzpicture}
    \node[fill=white,draw] (python) {\textcolor{gray}{[Python Output]}};
\end{tikzpicture}
\end{tcolorbox}

\begin{tcolorbox}[
    boxrule=0.5mm,
    arc=0.5mm,
    outer arc=0.5mm,
    colback=sm-back,
    colframe=sm-title,
    title=\#\#\# Summary
]
[Summary of the solution and key takeaways]
\end{tcolorbox}
\end{tcolorbox}
\caption{The universal text template designed for multi-paradigm reasoning. The contents for each paradigm have been omitted.}
\label{fig:universal-template}
\end{figure*}

As shown in~\cref{fig:universal-template}, we apply the designed universal text template on all training samples.

\subsection{Example Prompts for Dataset Enhancement}
\label{ass:enhance-prompt}

To guide LLMs in generating and refining reasoning paradigms during dataset enhancement, we developed specific prompts ($p_s$) tailored to each seed dataset. 
These prompts provide structured instructions for the models to follow, ensuring the generation of high-quality and logically consistent reasoning paradigms.

\cref{fig:enhance-prompt-tir} and~\cref{fig:enhance-prompt-lean} provide examples of such prompts designed for augmenting and refining formal proofs within the Lean 4 theorem prover for the Numina-TIR and Lean-Workbook datasets, respectively. 
The prompt specifies the input and output format and includes placeholders for the problem statement, informal proof, and corresponding formal proof in Lean 4. 
It also incorporates $N_{\text{shot}}$ few-shot examples to further guide the model. 
In our experiments, we follow the common settings~\citep{DBLP:journals/corr/abs-2310-06825@mistral-7b,DBLP:journals/corr/abs-2409-12122@qwen25-math,wang2025datawhisperer,zhao2024large} and set $N_{\text{shot}} = 8$.
This structured approach and setting are applied to all datasets to ensure comprehensive coverage and logical consistency across the enhanced dataset.

\subsection{Details of Training Settings}
\label{ass:training-settings}

In all stages of the PPT method, a learning rate of $2e-5$ and a warm-up ratio of $1\%$ are implemented. 
To enhance computational efficiency, the training process is conducted using distributed optimization with DeepSpeed ZeRO Stage 3~\citep{DBLP:conf/sc/RajbhandariRRH20@deepspeed} and is combined with Flash-Attention~\citep{DBLP:conf/iclr/Dao24@flash-2}.
Stage \ding{192} comprises $3$ epochs, whereas the stage \ding{193} and \ding{194} are conducted over $4$ epochs. 
A sequence length of $2,048$ tokens is used in Stages \ding{192} and \ding{193}, which is increased to $4,096$ tokens in stage \ding{194} to support more complex reasoning tasks.
Furthermore, we employ an annealing strategy at the end of stage \ding{194} with high-quality \datasetname samples, with the aim of further enhancing model accuracy in complex reasoning tasks.

\subsection{Benchmarks}
\label{ass:benchmarks}

We report the statistics of the evaluation datasets in~\cref{table:eval-statistics}.

\begin{table}[!tp]
\small
\centering
\begin{tabular}{l|r|r}
\toprule
 Dataset & \# of Test set & Avg. Length \\ 
\midrule
MATH  & 5,000 & 30.7 \\
GSM8K & 1,319 & 46.3 \\ 
AIME2024 & 30 & 59.2 \\
AMC2023 & 40 & 47.2 \\ 
\midrule
MiniF2f & 244 & 30.5 \\
\bottomrule
\end{tabular}
\caption{The statistics of the evaluation datasets.}
\label{table:eval-statistics}
\end{table}

\subsection{Details of Metrics}
\label{ass:details_metrics}

This study employs the pass@$N$ metric on the miniF2F benchmark, with the evaluation grounded in a sample budget of $N$. 
To ensure a fair comparison of computational cost across different generation schemes, this paper defines the sample budget according to the following rules.
\begin{itemize}
\setlength{\itemsep}{1pt}
\setlength{\parskip}{2pt}
\setlength{\parsep}{0pt}
\item For single-pass sampling methods, we define the sample budget $N$ as the total number of proofs generated, with large values of $N$ factorized for ease of comparison to tree search methods.
\item For best-first-search methods, following the notation in Llemma~\citep{DBLP:conf/iclr/AzerbayevSPSMJD24@llemma}, we present $N = N_{\text{att}} \times N_{\text{tact}} \times N_{\text{iter}}$ where $N_{\text{att}}$ denotes the number of best-first-search attempts, $N_{\text{tact}}$ denotes the number of tactics generated for each expansion, and $N_{\text{iter}}$ denotes the number of expansion iterations.
\item For tree-based search methods, e.g., RMaxTS~\citep{DBLP:journals/corr/abs-2408-08152@deepseek-prover-v1.5} and HTPS~\citep{DBLP:conf/nips/LampleLLRHLEM22@hypertree}, we present $N = N_{\text{att}} \times N_{\text{ex}}$, where $N_{\text{att}}$ denotes the number of tree search attempts, and $N_{\text{ex}}$ denotes the number of model generations invoked in tree expansions.
\item For our SMPS, we define the sample budget as $N = N_{\text{NLR}} \times N_{\text{SR}}$, where $N_{\mathrm{NLR}}$ denotes the number of initial semantic reasoning paths, and $N_{\mathrm{SR}}$ denotes the number of symbolic reasoning paths extended for each other reasoning paradigm.
\end{itemize}

\vspace{-1ex}
\begin{table*}[tp]
\small
\centering
\begin{adjustbox}{max width=0.8\textwidth}
\begin{tabular}{lrccc}
\toprule
\multirow{2.5}{*}{Model} & \multirow{2.5}{*}{SFT Data Size (k)} & MATH & GSM8k & \multirow{2.5}{*}{Average}  \\
\cmidrule{3-4}
 & & ZS Pass@1 & ZS Pass@1 &  \\
\midrule
WizardMath & 868 & 10.7 & 54.9 & 32.8 \\
MetaMath-7B & 2,790 & 19.8 & 66.5 & 43.2 \\
MetaMath-Llemma-7B & 2,790 & 30.0 & 69.2 & 49.6 \\
MetaMath-Mistral-7B & 2,790 & 28.2 & 77.7 & 53.0 \\
ToRA & 85 & 40.1 & 68.8 & 54.5 \\
ToRA-CODE & 85 & 44.6 & 72.6 & 58.6 \\
NuminaMath-7B-CoT & 859 & $54.4^{\dagger}$ & $66.6^{\dagger}$ & 60.5 \\
Xwin-Math-7B & 1,440 & 40.6 & 82.6 & 61.6 \\
DeepSeekMath-Instruct-7B & 776 & 46.8 & 73.6 & 64.2 \\
NuminaMath-7B-TIR & 931 & $55.3^{\dagger}$ & $73.6^{\dagger}$ & 64.5 \\
DeepSeekMath-RL-7B & 920 & 51.7 & 88.2 & 70.0 \\
DART-Math-7B & 1,175 & 53.6 & 86.6 & 70.1 \\
Qwen2.5-Math-7B-Instruct & 3,026 & \textbf{83.6} & \textbf{95.2} & \textbf{89.4} \\
\midrule
\modelname & 1,098 & {\ul 66.7} & {\ul 88.7} & {\ul 77.7} \\
\bottomrule
\end{tabular}
\end{adjustbox}
\caption{Zero-shot (ZS) performance on the MATH and GSM8K benchmarks for the arithmetic task, with data size in the Supervised Fine-Tuning (SFT) stage. $\dagger$ denotes our reported results with the open-sourced model weights. The best results are in \textbf{bold}, and the second-best results are {\ul underlined}.}
\label{table:full-arithmetic-res}
\end{table*}

\begin{figure*}[!tp]
\small
\begin{tcolorbox}[
    colback=white, 
    colframe=pink2, 
    boxrule=0.5mm,
    arc=0.5mm,
    outer arc=0.5mm,
    title=Example Prompts for TIR Dataset Enhancement
]
\textbf{System Prompt:}\\
You are an expert in Lean 4. Please respond to a math problem by translating the provided informal proof into Lean 4 code. Follow the format provided in the prompt. Please note that the informal proof and the formal proof need to be identical. Follow the format provided in the prompt.\\
\textbf{User Input:}\\
Now please translate the formal solution in Lean 4 following the instructions below. Please write the corresponding solution in Lean 4 (indicated by ``Formal proof in Lean 4: '') given the ``\# Problem: '' and ``\# Informal proof: '', filling in the ``\# Formal proof in Lean 4: '' section.\\
\\
You must respond in the following format: \\
\# Problem:  ...\\
\# Informal proof: ...\\
\# Formal proof in Lean 4: \\
\begin{verbatim}
```lean4
(lean 4 code for proving)
...

```
\end{verbatim}
Here are examples you may refer to:\\
\\
---\\
\colorbox{pink2!50}{\texttt{$\mathrm{N}$ few-shot examples}}\\
---\\
\# Problem: \colorbox{pink2!50}{\texttt{\{problem\}}}\\
\# Informal proof: \colorbox{pink2!50}{\texttt{\{informal\_proof\}}}\\
\# Formal proof in Lean 4: \\
\begin{verbatim}
```lean4
...

```
\end{verbatim}
\end{tcolorbox}
\caption{Example prompts for Numina-TIR dataset enhancement. The few-shot examples provided for the LLM have been omitted.}
\label{fig:enhance-prompt-tir}
\end{figure*}
\begin{figure*}[!tp]
\small
\begin{tcolorbox}[
    colback=white, 
    colframe=pink2, 
    boxrule=0.5mm,
    arc=0.5mm,
    outer arc=0.5mm,
    title=Example Prompts for Lean-Workbook Dataset Enhancement
]
\textbf{System Prompt:}\\
You are an expert in Lean 4 and formal mathematics. Your task is to explain how to construct formal proofs using Lean 4 syntax and tactics. Focus on the Lean 4 approach to theorem proving, rather than traditional mathematical reasoning.\\
\textbf{User Input:}\\
Please follow the instructions below to convert the Lean 4 code  (indicated by ``Formal proof in Lean 4: '')  into its informal proof, using the informal problem (indicated by ``Problem: '') as a guide. Please write the corresponding informal solution in natural language (indicated by ``Informal proof: '') given the ``\# Problem: '' and ``\# Formal proof in Lean 4: '', filling in the ``\# Informal proof: '' section.\\
<Instruction>\\
Analyze the given mathematical theorem and the corresponding Lean 4 code. Provide a detailed explanation of the proof process, adhering to the following guidelines:\\
1. Theorem structure: Clearly state the theorem, including its assumptions and conclusion.\\
2. Proof strategy: Explain the overall strategy employed in the proof, focusing on logical reasoning and mathematical deduction rather than calculation.\\
3. Step-by-step reasoning: Provide a detailed, step-by-step explanation of the proof process, ensuring that each step corresponds to an element in the Lean 4 code.\\
4. Logical deduction: Emphasize how each step of the proof follows logically from the previous steps or from the given assumptions.\\
5. Mathematical concepts: Discuss any specific mathematical concepts, notations, or definitions used in the proof, such as divisibility or exponentiation.\\
6. Abstraction: Present the proof in a general, abstract form that could be applied to similar problems, rather than focusing on specific numerical calculations.\\
7. Correspondence to code: Ensure that the logical flow of your proof explanation aligns with the structure and tactics used in the Lean 4 code, without explicitly mentioning Lean-specific terminology.\\
Avoid using syntax or terminology specific to formal proof systems. Instead, focus on presenting a rigorous mathematical argument using general logical principles and mathematical language. The proof should follow the reasoning path implied by the Lean 4 code but be accessible to readers unfamiliar with formal proof assistants.\\
</Instruction>\\
You must respond in the following format:\\
\# Problem:  ...\\
\# Tags:  ...\\
\# Formal proof in Lean 4: \\
\begin{verbatim}
```lean4
(lean 4 code for proving)

```

# Informal proof: 
(Informal reasoning path for proving the problem)
\end{verbatim}
Here are examples you may refer to:\\
---\\
\colorbox{pink2!50}{\texttt{$\mathrm{N}$ few-shot examples}}\\
---\\
\# Problem: \colorbox{pink2!50}{\texttt{\{problem\}}}\\
\# Tags: \colorbox{pink2!50}{\texttt{\{tag\}}}\\
\# Formal proof in Lean 4: \\
\begin{verbatim}
```lean4
\end{verbatim}
\colorbox{pink2!50}{\texttt{\{lean\_workbook code\}}}\\
\begin{verbatim}
```
\end{verbatim}
\# Informal proof: ...
\end{tcolorbox}
\caption{Example pompts for Lean-Workbook dataset enhancement. The few-shot examples provided for the LLM have been omitted.}
\label{fig:enhance-prompt-lean}
\end{figure*}

\subsection{Experimental Details of MATH and GSM8K benchmarks}
\label{ass:details-arithmetic}

For arithmetic-related results, we analyze the number of mathematical samples utilized by the baseline model following the pre-training phase, as illustrated in~\cref{fig:intro-result}. Specifically, as shown in~\cref{table:full-arithmetic-res}, we present the sample quantity along with the model performance.

\section{Additional Analysis}

\subsection{The Risk of Data Leakage}
\label{ass:data-leak}
We measure the Levenshtein distance~\citep{10.5555/1822502@levenshtein} between problems in the training dataset and those in the test dataset to mitigate the risk of data leakage. 
During the experiments, we set a similarity threshold of $0.7$ and excluded cases from the training dataset that exceeded this threshold.
As a result, approximately $1,000$ cases, accounting for less than $0.6\%$ of our training dataset, were removed to prevent potential leakage.

\begin{table}[!tp]
\small
\centering
\begin{adjustbox}{max width=0.48\textwidth}
\begin{tabular}{lcccc}
\toprule
\multirow{2}{*}{Model} & \multicolumn{2}{c}{MATH} & \multicolumn{2}{c}{GSM8k} \\
\cmidrule{2-5}
 & Pass@1 & Maj@8  & Pass@1 & Maj@8 \\
\midrule
Llama-3.1-8B & 58.18 & 59.46 & 84.00 & 87.26 \\
DeepSeekMath-Base-7B & 66.74 & 71.70 & 88.70 & 91.43 \\
\bottomrule
\end{tabular}
\end{adjustbox}
\caption{Zero-shot results of different evaluation strategies across MATH and GSM8k benchmarks.}
\label{table:voting-res}
\end{table}

\subsection{Different Evaluation Strategies on Arithmetic Benchmarks}
\label{ass:diff-eval-strategies}

\cref{table:voting-res} explores the impact of various evaluation strategies on the CoR framework. 
The majority vote strategy benefits the CoR framework. 
Specifically, a majority vote strategy with $8$ candidate samples on GSM8K can exceed GPT-4o's Pass@1 performance ($90.5\%$).

\subsection{Ethical Considerations}

In this work, we present a method that achieves robust zero-shot multi-paradigm reasoning capabilities. 
However, this comes with potential risks, which include two primary social and ethical concerns: 1) the risk of tool misuse, and 2) pre-existing risks embedded in the backbone model parameters.
The CoR framework supports various reasoning paradigms, some of which are deeply integrated with specific tools. 
For instance, the AR paradigm may involve unvetted Python code libraries. 
Therefore, we recommend performing pre-deployment audits on tools that may be utilized by the model. 
Furthermore, our method is compatible with existing LLMs, and several studies have highlighted the presence of biased behaviors in models like LLaMa~\citep{DBLP:journals/corr/abs-2410-20739@bais}. 
To mitigate ethical risks, we encourage the use of risk-free language models during deployment, which can reduce potential harm. 
Recent research suggests incorporating ethical alignment processes, such as filtering training data or generated content, to minimize unnecessary risks~\cite{DBLP:conf/sigir-ap/YuWZZYS23@ealm}. 
Overall, we advocate for open discussions regarding the use of our framework, as a transparent and public environment can reduce the likelihood of misuse.

\section{Case Studies}
\label{ass:case_study}

\subsection{Cases of Instruction-free Reasoning}
\label{ass:instruction-free-reasoning-case}

In the absence of explicit instructions, as illustrated in~\cref{fig:instruction-free-reasoning-case}, the model generally adopts a paradigm sequence that aligns with the task types present in the training data, as described by MPM. 
The model transitions between paradigms autonomously. 
For example, during a computational task, the model initiates reasoning with the paradigm NLR, proceeds to SR for logical verification, and ultimately employs AR to achieve enhanced computational performance.

\subsection{Qualitative Analysis of Error Cases}
\label{ass:error-case}
We now present a preliminary error pattern analysis for \modelname across two tasks: arithmetic computation and theorem proving.

In arithmetic computation, an analysis of 200 randomly sampled error cases reveals four primary types of errors: comprehension errors or incorrect premise assumptions (33 cases), logical fallacies (26 cases), computational errors (56 cases), and incomplete proofs or code syntax errors, which were most prevalent in the AR/SR paradigms (85 cases). Notably, cross-paradigm self-correction behaviors were observed. For example, AR rectified erroneous calculations initially produced by NLR, and SR supplemented incomplete logical steps that were missing in NLR outputs. In theorem proving, a separate analysis of 50 randomly sampled error cases revealed the same four error types, with the following distribution: 2 comprehension errors, 6 logical fallacies, 9 computational errors, and 33 incomplete proofs or syntax errors. Cross-paradigm self-correction behaviors were again consistently observed. As shown in ~\cref{fig:error-case} and ~\cref{fig:self_correction_case}, we provide an example of error cases and an example of self-correction behavior. 

We attribute these errors to the following potential causes. Despite MPM’s systematic integration of multi-paradigm reasoning paths and PPT’s paradigm-progressive training as discussed in ~\cref{ss:collecting_dataset} and ~\cref{ss:inference} to mitigate distribution shifts, the overwhelming prevalence of NLR examples in the pretraining data \textemdash particularly the scarcity of SR-specific instances \textemdash has resulted in persistent competency gaps across paradigms. Additionally, CoR inherits the base LLM’s fixed-context window, which introduces the risk of truncating long reasoning chains or omitting key steps, leading to incomplete outputs. Furthermore, the observed self-correction capabilities can be attributed to the complementarity of reasoning paradigms within CoR, where one reasoning paradigm can rectify errors made by another, thus validating the effectiveness of our multi-paradigm approach. Additionally, feedback from multi-step reasoning validation allows subsequent paradigms to verify earlier outputs, facilitating error detection and correction throughout the reasoning process.

\begin{figure*}[!tp]
\small
\begin{tcolorbox}[
    colback=white, 
    colframe=ar-back, 
    boxrule=0.5mm,
    arc=0.5mm,
    outer arc=0.5mm,
    title=Examples of Instruction-free Reasoning Generated by CoR
]
\textbf{Problem:}\\
Cities A and B are 45 miles apart. Alicia lives in A and Beth lives in B. Alicia bikes towards B at 18 miles per hour. Leaving at the same time, Beth bikes toward A at 12 miles per hour. How many miles from City A will they be when they meet?\\

\textbf{Solution:}\\
Let's go through this step-by-step in natural language.  To determine how many miles from City A Alicia and Beth will meet, we can use the concept of relative speed and the distance between their homes.\\
1. Calculate the relative speed between Alicia and Beth:\\
- Alicia's speed: 18 miles per hour\\
- Beth's speed: 12 miles per hour\\
- Relative speed = 18 + 12 miles per hour\\
2. Calculate the time it takes for them to meet:\\
- Distance between City A and City B: 45 miles\\
- \( \text{Time to meet} = \text{Distance} \div \text{Relative speed} \)\\
3. Calculate the distance each person covers before meeting:\\
- \( \text{Distance covered by Alicia} = \text{Speed of Alicia} \times \text{Time to meet} \)\\
- \( \text{Distance covered by Beth} = \text{Speed of Beth} \times \text{Time to meet} \).\\

Next, let's write the corresponding formal proof in Lean 4 to prove this.\\
\# Formal proof in Lean 4: \\
\setlength{\topsep}{-1mm} 
\setlength{\partopsep}{-1mm} 
\begin{verbatim}
```lean4
def distance_AB := 45  -- miles
def speed_Alicia := 18  -- miles per hour
def speed_Beth := 12  -- miles per hour
-- Calculate the relative speed
def relative_speed := speed_Alicia + speed_Beth  -- miles per hour
-- Calculate the time to meet
def time_to_meet := distance_AB.toFloat / relative_speed.toFloat  -- in hours
-- Calculate the distance each person covers before meeting
def distance_covered_by_Alicia := speed_Alicia.toFloat * time_to_meet
def distance_covered_by_Beth := speed_Beth.toFloat * time_to_meet
-- Calculate how many miles from City A they will be when they meet
def distance_from_A_when_meet := distance_covered_by_Alicia
#eval distance_from_A_when_meet  -- returns 27.0
```
```lean_output
27.000000
```
\end{verbatim}

Let's use Python code to perform these calculations.\\
\# Code in Python: \\
\begin{verbatim}
```python
import sympy as sp
# Given values
distance_AB = 45  # miles
speed_Alicia = 18  # miles per hour
speed_Beth = 12  # miles per hour
# Calculate the relative speed
relative_speed = speed_Alicia + speed_Beth  # miles per hour
# Calculate the time to meet
time_to_meet = distance_AB / relative_speed  # in hours
# Calculate the distance each person covers before meeting
distance_covered_by_Alicia = speed_Alicia * time_to_meet
distance_covered_by_Beth = speed_Beth * time_to_meet
# Calculate how many miles from City A they will be when they meet
distance_from_A_when_meet = distance_covered_by_Alicia
print(distance_from_A_when_meet)
```
```output
27.0
```
\end{verbatim}

Alicia and Beth will meet boxed{27} miles from City A.

\end{tcolorbox}
\caption{A case of the reasoning trajectory produced by CoR in an instruction-free setting.}
\label{fig:instruction-free-reasoning-case}
\end{figure*}
\begin{figure*}[!tp]
\small
\begin{tcolorbox}[
    colback=white, 
    colframe=black!50, 
    boxrule=0.5mm,
    arc=0.5mm,
    outer arc=0.5mm,
    title=Examples of Incomplete Proofs or Code Syntax Errors
]
\textbf{Problem:}\\
What is the value of \( \sqrt{1{,}\!000{,}\!000} - \sqrt[3]{1{,}\!000{,}\!000} \)? Show that it is 900.\\

\textbf{Solution:}\\
Let's go through this step-by-step in natural language. \\
We need to prove the following theorem:

\begin{verbatim}
theorem mathd_algebra_208 : Real.sqrt 1000000 - 1000000 ^ ((1 : ℝ) / 3) = 900 := by
\end{verbatim}

We have the following:

\begin{align*}
\sqrt{1,000,000} - \sqrt[3]{1,000,000} &= \sqrt{10^6} - \sqrt[3]{10^6} \\
&= (10^6)^{\frac{1}{2}} - (10^6)^{\frac{1}{3}} \\
&= 10^{6 \cdot \frac{1}{2}} - 10^{6 \cdot \frac{1}{3}} \\
&= 10^3 - 10^2 = 1000 - 100 = 900.
\end{align*}

So, the value of \( \sqrt{1,000,000} - \sqrt[3]{1,000,000} \) is indeed 900.

Next, let's write the corresponding formal proof in Lean 4 to prove this.\\
\# Formal proof in Lean 4: \\
\setlength{\topsep}{-1mm}
\setlength{\partopsep}{-1mm} 
\begin{verbatim}
```lean4
theorem mathd_algebra_208 : Real.sqrt 1000000 - 1000000 ^ ((1 : ℝ) / 3) = 900 := by
norm_num [sqrt_eq_iff_mul_self_eq_of_pos, pow_one]
```
```lean_output
Execution Failed.
```
\end{verbatim}

The original code has a syntax error. The proof block after := by is incomplete. Proof failed.

\end{tcolorbox}
\caption{A proof task failure case caused by incomplete proofs or code syntax errors.}
\label{fig:error-case}
\end{figure*}
\begin{figure*}[!tp]
\small
\begin{tcolorbox}[
    colback=white, 
    colframe=black!50, 
    boxrule=0.5mm,
    arc=0.5mm,
    outer arc=0.5mm,
    title=Examples of Instruction-free Reasoning Generated by CoR
]
\textbf{Problem:}\\
Let \( b \ge 2 \) be an integer. Call a positive integer \( n \) \textit{\( b \)-eautiful} if it has exactly two digits when expressed in base \( b \), and these two digits sum to \( \sqrt{n} \). \\
For example, \( 81 \) is \( 13 \)-\textit{eautiful} because \( 81 = \underline{6}\,\underline{3}_{13} \) and \( 6 + 3 = \sqrt{81} \).\\
Find the least integer \( b \ge 2 \) for which there are more than ten \( b \)-\textit{eautiful} integers.\\

\textbf{Solution:}\\
Let's go through this step-by-step in natural language. To solve this problem, we need to find the smallest integer \( b \ge 2 \) such that there are more than ten \( b \)-beautiful integers.\\
1. Understanding the Problem:\\
- We need to identify \( b \)-beautiful integers, which have exactly two digits in base \( b \).\\
- These integers \( n \) must satisfy the condition that the sum of their digits equals \( \sqrt{n} \).\\
...\\
6. Conclusion:\\
- The least integer \( b \ge 2 \) for which there are more than ten \( b \)-beautiful integers is \( b = 13 \).\\
Therefore, the smallest \( b \) such that there are more than ten \( b \)-beautiful integers is \( b = 13 \).\\

Next, let's write the corresponding formal proof in Lean 4 to prove this.\\
\# Formal proof in Lean 4: \\
\setlength{\topsep}{-1mm}
\setlength{\partopsep}{-1mm}
\begin{verbatim}
```lean4
theorem lean_workbook_plus_1377 (b : ℕ) (hb : 2 ≤ b) 
(h : (∃ n, (Nat.sqrt n = √n ∧ Nat.digits b n = [n / b, n % b])) 
  → ¬ (∃ n, (Nat.sqrt n = √n ∧ Nat.digits b n = [n / b, n % b]))) : 13 ≤ b := 
  by
  contrapose! h
  interval_cases b <;> norm_num [Nat.digits_eq_of_eq_mul_left, Nat.digits_eq_of_eq_mul_right]
```
```lean_output
27.000000
```
\end{verbatim}

Let's use Python code to perform these calculations.\\
\# Code in Python: \\
\begin{verbatim}
```python
import sympy as sp
import math
def is_b_beautiful(n, b):
\"\"\"Check if a number n is b-beautiful.\"\"\"
digits = [n // b, n % b]
return sum(digits) == math.sqrt(n)
def count_b_beautiful_numbers(b):
\"\"\"Count the number of b-beautiful numbers for a given b.\"\"\"
count = 0
for n in range(1, b * b):
if is_b_beautiful(n, b):
count += 1
return count
def find_smallest_b_with_more_than_ten_beautiful_numbers():
\"\"\"Find the smallest b >= 2 such that there are more than ten b-beautiful numbers.\"\"\"
b = 2
while True:
if count_b_beautiful_numbers(b) > 10:
return b
b += 1
# Find the smallest b
smallest_b = find_smallest_b_with_more_than_ten_beautiful_numbers()
print(smallest_b)
```
```output
211
```
\end{verbatim}

The smallest integer \( b \ge 2 \) for which there are more than ten \( b \)-beautiful integers is \( b = 211 \). Thus, the answer is \( \boxed{211} \).

\end{tcolorbox}
\caption{A case of self-correction behaviors.
Some lengthy content is omitted with "..." for brevity.}
\label{fig:self_correction_case}
\end{figure*}

\end{document}